\makeatletter
\PassOptionsToPackage{hyphenbreaks}{breakurl}
\PassOptionsToPackage{table}{xcolor}

\@namedef{ver@breakurl.sty}{9999/12/31}

\newcommand{\breakurldefns}{} 
\newcommand{\burl}[1]{\url{#1}}
\makeatother

\documentclass[iicol]{sn-jnl}

\usepackage{algorithm}
\usepackage{algpseudocode}
\usepackage{amsmath}
\usepackage{adjustbox}
\usepackage{makecell}
\usepackage{chngcntr}
\usepackage{needspace}
\usepackage{hyperref}
\usepackage{array}
\usepackage{url} 
\usepackage{graphicx}
\usepackage{epstopdf}
\usepackage{graphicx}%
\usepackage{multirow}%
\usepackage{amsmath,amssymb,amsfonts}%
\usepackage{amsthm}%
\usepackage{mathrsfs}%
\usepackage[title]{appendix}%
\usepackage{xcolor}%
\usepackage{textcomp}%
\usepackage{manyfoot}%
\usepackage{booktabs}%
\usepackage{algorithm}%
\usepackage{algorithmicx}%
\usepackage{algpseudocode}%
\usepackage{listings}%
\usepackage{graphicx}
\usepackage{subcaption}
\usepackage{placeins}
\usepackage{dsfont}

\definecolor{bestgray}{gray}{0.9} 

\theoremstyle{thmstyleone}%
\theoremstyle{thmstyletwo}%

\theoremstyle{thmstylethree}%

\usepackage[backend=bibtex, sorting=none]{biblatex}
\begin{document}

\title[Article Title]{Inconsistency Masks: Harnessing Model Disagreement for Stable Semi-Supervised Segmentation}

\author*[1]{\fnm{Michael R. H.} \sur{Vorndran}}\email{michiv2012@gmail.com}

\author[2,3]{\fnm{Bernhard F.} \sur{Roeck}}\email{broeck@uni-koeln.de}

\affil*[1]{Independent Researcher} 

\affil[2]{\orgdiv{Institute for Genetics}, \orgname{University of Cologne}, \orgaddress{\street{Joseph-Stelzmann-Straße 26}, \city{Cologne}, \postcode{50931}, \country{Germany}}}

\affil[3]{\orgdiv{CECAD Cluster of Excellence}, \orgname{University of Cologne}, \orgaddress{\street{Joseph-Stelzmann-Straße 26}, \city{Cologne}, \postcode{50931}, \country{Germany}}}


\abstract{A primary challenge in semi-supervised learning (SSL) for segmentation is the confirmation bias from noisy pseudo-labels, which destabilizes training and degrades performance. We propose Inconsistency Masks (IM), a framework that reframes model disagreement not as noise to be averaged away, but as a valuable signal for identifying uncertainty. IM leverages an ensemble of teacher models to generate a mask that explicitly delineates regions where predictions diverge. By filtering these inconsistent areas from input-pseudo-label pairs, our method effectively mitigates the cycle of error propagation common in both continuous and iterative self-training paradigms. Extensive experiments on the Cityscapes benchmark demonstrate IM's effectiveness as a general enhancement framework: when paired with leading approaches like iMAS, U$^2$PL, and UniMatch, our method consistently boosts accuracy, achieving superior benchmarks across ResNet-50 and DINOv2 backbones, and even improving distilled architectures like SegKC. Furthermore, the method's robustness is confirmed in resource-constrained scenarios where pre-trained weights are unavailable. On three additional diverse datasets from medical and underwater domains trained entirely from scratch, IM significantly outperforms standard SSL baselines. Notably, the IM framework is dataset-agnostic, seamlessly handling binary, multi-class, and complex multi-label tasks by operating on discretized predictions. By prioritizing training stability, IM offers a generalizable and robust solution for semi-supervised segmentation, particularly in specialized areas lacking large-scale pre-training data. 

The full code is available at: 
\href{https://github.com/MichaelVorndran/InconsistencyMasks}{https://github.com/MichaelVorndran/InconsistencyMasks}}




\maketitle

\section{Introduction}\label{sec1}

In the rapidly evolving field of computer vision, semantic segmentation plays a pivotal role in understanding and interpreting visual information ~\cite{long_fully_2014, minaee2021image}. However, a significant and persistent challenge in this domain is the scarcity of high-quality labeled datasets, especially in niche or emerging areas ~\cite{tajbakhsh2020embracing}. Despite the growing number of publicly accessible datasets and the development of better annotation tools like Meta’s SAM~\cite{kirillov_segment_2023}, the development of specialized applications still often requires extensive, pixel-perfect annotation, a process that is both time-consuming and expensive~\cite{cordts_cityscapes_2016}. Semi-supervised learning (SSL) offers a promising path forward, aiming to leverage large quantities of readily available unlabeled data alongside a small set of labeled examples~\cite{chapelle_semi-supervised_2009, zhu_introduction_2009}.

However, the pseudo-labeling paradigm central to many SSL methods~\cite{lee_pseudo-label_2013} introduces a fundamental risk: training instability driven by confirmation bias~\cite{arazo2020pseudo}. Whether in a continuous loop where a model learns from its own predictions in real-time (e.g., FixMatch \cite{sohn2020fixmatch}), or in an iterative framework where a student learns from a teacher's fixed predictions~\cite{xie_self-training_2019, yang2022st++}, the core problem persists: if the pseudo-labels are noisy, the model will learn to amplify its own errors. As our experiments show, this can manifest as catastrophic performance collapse in later training epochs, even for state-of-the-art methods. Preventing this cycle of error propagation, where a model becomes increasingly confident in its own mistakes, is a key challenge for achieving stable and reliable semi-supervised learning~\cite{u2pl, iMAS, oliver2018realistic}.

The landscape of computer vision is currently being reshaped by large-scale foundation models. Recent works like UniMatch~v2 ~\cite{UniMatch_v2} have achieved remarkable results on standard benchmarks by fine-tuning massive, self-supervised backbones like DINOv2 ~\cite{oquab_dinov2_2023}. This paradigm, which leverages a model's prior knowledge from vast, general-purpose datasets, represents the state-of-the-art when significant computational resources are available. While our primary motivation is to enable robust learning in resource-constrained, niche domains where foundation models may not apply due to domain shift or hardware limitations, we discover that the stability provided by Inconsistency Masks is consistently beneficial across diverse regimes. Our experiments demonstrate that IM addresses the challenge of confirmation bias from residual label noise, which persists even in large-scale foundation models. Consequently, our framework not only enables training from scratch but also acts as a synergistic booster, pushing state-of-the-art methods like UniMatch~v2 and SegKC ~\cite{segkc} to new heights. This makes it uniquely applicable to the long tail of niche domains where custom models must be built from limited data.

Our focus on these resource-constrained scenarios is motivated by a direct challenge in high-throughput microscopy, which led to the development of our open-source CellLocator application ~\cite{vorndran_celllocator_2024}, a tool already in use in published research ~\cite{ujevic_tbk1_2024, roeck_ferroptosis_2025}. Analyzing live-cell imaging presents a significant hurdle: a single field of view can contain thousands of cells, making traditional instance segmentation computationally prohibitive on typical workstation or laptop hardware and unreliable in dense cell cultures. Furthermore, determining cell viability (alive/dead) often requires dedicating a precious fluorescence channel to a death marker, limiting experimental capacity. To solve these issues, we developed a label-free, position-based approach. A model is trained to predict cell status and location from the brightfield image alone, freeing up all fluorescence channels for experimental markers. This, however, results in an unconventional multi-label segmentation task where the outputs are not mutually exclusive. The necessity of training a model for this unconventional task from a limited number of annotated images was the primary motivation for developing a more stable and dataset-agnostic SSL framework.

To address this, we propose Inconsistency Masks (IM), a novel framework that reframes model disagreement not as noise to be averaged away, but as a direct signal of uncertainty. We begin by training an ensemble of teacher models on the initial labeled data. Instead of creating a consensus pseudo-label by voting, we identify precisely where the models' predictions diverge. These regions of disagreement are consolidated into a binary "inconsistency mask." This mask is then used to filter out the uncertain areas from both the input image and the consensus pseudo-label. By explicitly preventing the student from training on ambiguous or contested pixels, IM effectively mitigates the cycle of error propagation and stabilizes the self-training process.

This paper makes the following contributions:
\begin{itemize}
\item We introduce Inconsistency Masks (IM), a simple, versatile, and stable framework for semi-supervised semantic segmentation that harnesses model disagreement to filter uncertainty.
\item We conduct a rigorous benchmark on the Cityscapes dataset~\cite{cordts_cityscapes_2016}, demonstrating that our framework achieves best-in-class results under a unified reproduction setting with ResNet-50 backbones and modern DINOv2 foundation models when applied to leading methods.
\item We demonstrate the robustness and generalizability of IM across four diverse datasets, including medical (ISIC 2018~\cite{codella_skin_2019, tschandl_ham10000_2018}, HeLa), underwater (SUIM~\cite{islam_semantic_2020}), and urban scenes (Cityscapes~\cite{cordts_cityscapes_2016}).
\item We release our full implementation and the annotated HeLa dataset, and highlight the framework's practical utility through its successful application in the open-source CellLocator tool~\cite{vorndran_celllocator_2024}.
\end{itemize}

\section{Related Work}\label{sec2}

Our work builds upon established principles in semi-supervised learning while introducing a distinct approach to handling uncertainty. We situate our method in the context of consistency regularization, pseudo-labeling, and ensemble techniques.

\subsection{Consistency Regularization and Pseudo-Labeling}\label{subsec2}

A cornerstone of modern semi-supervised learning is the consistency regularization principle, which posits that a model's prediction should remain stable under small perturbations of its input. This is often implemented by enforcing consistency between predictions on weakly and strongly augmented views of the same unlabeled image. FixMatch~\cite{sohn2020fixmatch} exemplifies this, using a model's confident prediction on a weakly augmented image as a pseudo-label to supervise its prediction on a strongly augmented version.

This concept of generating pseudo-labels is a dominant paradigm in SSL~\cite{lee_pseudo-label_2013}. The primary challenge, however, is the quality of these labels. Noisy or incorrect pseudo-labels can lead to confirmation bias, where the model becomes increasingly confident in its own errors~\cite{arazo2020pseudo}. To mitigate this, methods like CrossMatch~\cite{zhao2024crossmatch} introduce more complex perturbation strategies and knowledge distillation between different model views. Fuzzy Positive Learning (FPL)~\cite{fuzzypositivlearning} moves beyond hard labels, allowing a pixel to be associated with a small set of ``fuzzy positive'' classes to handle ambiguity at class boundaries. These methods refine the pseudo-labeling process by incorporating more sophisticated notions of consistency and label representation. While powerful, they operate under a continuous self-training loop where the student is constantly influenced by its own (or its EMA-teacher's) evolving predictions. In contrast, our IM framework uses discrete training generations and leverages the disagreement across a decoupled ensemble to explicitly filter out uncertain regions before they are presented to the student.

\subsection{Advanced SSL for Semantic Segmentation}\label{subsec3}

Recent methods have developed highly specialized strategies for segmentation. UPS (Uncertainty-aware Pseudo-label Selection)~\cite{UPS} utilizes Monte Carlo dropout to estimate model uncertainty, using it to select reliable pixels for pseudo-labeling. iMAS (Instance-Specific and Model-Adaptive Supervision)~\cite{iMAS} dynamically generates instance-specific soft labels and employs a model-adaptive training process with adaptive augmentations. U$^2$PL (Unreliable Pseudo-Labels)~\cite{u2pl} goes a step further by explicitly leveraging unreliable pixels through a contrastive learning objective, encouraging representations of unreliable pixels to be distinct from reliable class prototypes.
These state-of-the-art methods represent a shift towards more granular, pixel-level supervision and uncertainty modeling. While powerful, this added complexity -- involving contrastive learning heads, adaptive online augmentations, or multi-component loss functions -- presents practical challenges for rapid prototyping and adaptation to new problems. Our Inconsistency Mask approach offers a conceptually simpler alternative. Instead of modeling uncertainty implicitly through entropy or contrastive losses, we define it directly and explicitly as the disagreement between independent teacher models. This allows us to use standard segmentation losses and architectures, making IM a more straightforward and easily integrable framework for stabilizing training.

\subsection{SSL with Foundation Models}\label{subsec4}

A recent and powerful trend in SSL is the hybrid approach of combining semi-supervised fine-tuning with large-scale foundation models. Methods such as UniMatch v2~\cite{UniMatch_v2} leverage powerful backbones like DINOv2 \cite{oquab_dinov2_2023}, which have been pre-trained on massive, unlabeled datasets, to achieve state-of-the-art performance. These methods represent the upper echelon of performance when a relevant foundation model and significant computational resources are available. Our work investigates the fundamental stability of SSL mechanisms to bridge this gap. The proposed framework enables effective training from scratch on standard architectures while simultaneously acting as a synergistic booster for massive, pre-trained foundation models. We aim to provide an accessible and broadly applicable solution that is effective for building specialized models even when pre-trained backbones are unavailable.

\subsection{Ensemble Methods in Deep Learning}\label{subsec5}

The use of model ensembles to improve performance is a well-established technique~\cite{goos_ensemble_2000}. Methods like snapshot ensembles~\cite{huang2017snapshotensemblestrain1} or simple averaging of predictions from multiple independently trained models often yield a more robust final prediction than any single model. In SSL, ensembles are sometimes used to generate higher-quality pseudo-labels through majority voting or averaging soft predictions.
Our framework diverges fundamentally from this traditional use of ensembles. We do not use the ensemble's consensus prediction directly as the final output. Instead, we are primarily interested in the variance of the predictions, not the mean. The regions where the models in our ensemble disagree are precisely the areas we identify as uncertain. While we use the consensus prediction to form the basis of the pseudo-label, our key innovation is the creation and application of the Inconsistency Mask derived from this disagreement. This transforms the ensemble from a tool for final prediction into a mechanism for generating a data-filtering signal, thereby guiding the training of a new, more robust student. Crucially, the final inference is performed by this single student model, avoiding the computational overhead of running an ensemble at test time. As our analysis in Section~\ref{sec_analysis_robustness} and the results in Table~\ref{tab_Study_B} show, this approach of filtering via disagreement is substantially more effective than simply training on pseudo-labels generated by a standard ensemble vote.

\section{Methodology}\label{subsec_methodology}

Our approach is centered on a novel mechanism for filtering uncertainty called Inconsistency Masks (IM), which is integrated into an iterative, generational semi-supervised learning framework. This section first defines the creation and application of IM, then describes the overall training pipeline, and finally details the advanced hybrid variants we evaluate.

\subsection{Problem Formulation}
\label{sec:problem_formulation}

We consider semi-supervised segmentation with a small labeled set
$\mathcal{D}_L=\{(x_i,y_i)\}_{i=1}^{N_L}$ and a larger unlabeled set
$\mathcal{D}_U=\{u_j\}_{j=1}^{N_U}$, where $x,u\in\mathbb{R}^{H\times W\times C}$
are images and $y$ are pixel-wise annotations (e.g., class labels or binary masks).
Our goal is to learn a model $f_\theta$ that generalizes to the test distribution.

In our generational framework, we construct a teacher ensemble
$\mathcal{T}=\{f_{\theta_k}\}_{k=1}^K$ (derived from independent models or checkpoints)
to supervise a student model $f_{\theta_S}$.
For an unlabeled image $u\in\mathcal{D}_U$, each teacher produces a prediction map $\hat{y}_k = f_{\theta_k}(u)$ (discretized via argmax or thresholding).

\paragraph{Inconsistency Mask.}
We define the \textbf{Inconsistency Mask (IM)} $M\in\{0,1\}^{H\times W}$ as a binary indicator of uncertainty.
For any pixel location $p$, we flag disagreement if the teachers are not unanimous:
\begin{equation}
    M^{(p)} = \mathds{1}\left[ \exists\, k,l:\ \hat{y}_k^{(p)} \neq \hat{y}_l^{(p)} \right],
\end{equation}
where $M^{(p)}=1$ denotes inconsistency and $M^{(p)}=0$ denotes consensus.
In our experiments, we typically use $K=2$ (pairwise comparison), where this condition effectively identifies any divergence between the teachers. We selected $K=2$ based on an ablation sweep (see Appendix~\ref{Appendix_D_K_e_d}) showing that larger ensembles did not yield significant performance gains to justify the added computational cost.

\paragraph{Consensus Pseudo-Label.}
For pixels where the ensemble reaches consensus ($M^{(p)}=0$), the pseudo-label is unique ($\hat{y}_k^{(p)}$ is identical for all $k$). We define the filtered target $\tilde{y}$ as:
\begin{equation}
    \tilde{y}^{(p)} =
    \begin{cases}
        \hat{y}_1^{(p)}, & M^{(p)}=0,\\
        y_{\text{unc}}, & M^{(p)}=1,
    \end{cases}
\end{equation}
where $y_{\text{unc}}$ denotes the ignore index for multi-class tasks or the background class ($0$) for binary tasks. In Study~B, we implement $y_{\text{unc}}$ as an explicit uncertainty class (Class~0) by shifting all semantic class labels by one (see Appendix~\ref{sec:handling_inconsistent_pixels}), allowing the model to be trained conservatively on ambiguous regions.

\paragraph{Input Filtering.}
Our method additionally filters the \emph{input} space to mask ambiguous regions.
We mask discordant regions in the unlabeled image (broadcasting $M$ across all $C$ channels):
\begin{equation}
    \tilde{u} = u \odot (1 - M).
\end{equation}

\paragraph{Training Objective.}
The student is optimized using supervised learning on labeled data and masked pseudo-supervision on unlabeled data. We sum the unsupervised loss over all pixels with non-ignored targets (i.e., $\tilde{y}^{(p)}\neq \texttt{ignore}$):
\begin{equation}
\begin{split}
    \mathcal{L}(\theta_S) &= \mathcal{L}_{sup}(\mathcal{D}_L) \\
    &\quad + \lambda \sum_{\substack{u\in\mathcal{D}_U \\ p:\tilde{y}^{(p)}\neq \texttt{ignore}}} \ell_{\text{task}}\!\left(f_{\theta_S}(\tilde{u})^{(p)},\, \tilde{y}^{(p)}\right),
\end{split}
\end{equation}

where $\ell_{\text{task}}$ is the task-specific loss (e.g., Cross-Entropy for classification, or MSE for regression-style soft targets).

\subsection{Inconsistency Masks: Harnessing Model Disagreement}\label{subsubsec_im}

Traditional ensemble methods in SSL aim to create a superior pseudo-label by averaging the predictions of multiple models, effectively discarding the information contained in their disagreements. In contrast, our approach elevates this disagreement to be a primary signal of uncertainty. While soft-voting variance could provide a continuous measure of uncertainty, it often necessitates finding task-specific thresholds. We instead do not rely on any additional confidence thresholding beyond standard hard discretization, yielding a confidence-threshold-free mechanism that is robust across diverse regimes--from training huge foundation models to learning from scratch--while aggressively filtering boundary ambiguity.
Given a set of $K \geq 2$ teacher models, we generate predictions for each unlabeled image. For each pixel, we then determine if all $K$ models agree on the class assignment. The output of this process is twofold: a high-confidence consensus pseudo-label and an Inconsistency Mask (IM).

\subsection{Binary and Multi-Class IM Generation}\label{subsubsec_im_def}

For binary segmentation tasks, the process is defined in Algorithm~\ref{alg_binary_im}. The hard predictions (thresholded at $0.5$) from each of the $K$ models are stacked.
\begin{itemize}
\item The final pseudo-label is formed from pixels where all models agree on the foreground class. All other pixels, including those where models disagree or agree on the background, are assigned to class 0 (which serves as the background or explicit uncertainty class, see Appendix~\ref{sec:handling_inconsistent_pixels}).
\item The Inconsistency Mask (IM) is a binary mask where a pixel is marked as $1$ if there is any disagreement among the models (i.e., the sum of votes is not $0$ or $K$), and $0$ otherwise.
\end{itemize}

\begin{algorithm*}
\caption{Binary Predictions to Inconsistency Masks and Consensus Pseudo-Label}
\label{alg_binary_im}
\begin{algorithmic}[1]
\Require A non-empty list of prediction masks $pm$, with the number of $pm \geq 2$
\Function{PredMasksToImBinary}{$pm$}
    \State $P \gets$ stack the matrices in $pm$ along a new first axis
    \State $S \gets$ sum $P$ along the first axis
    \State $K \gets$ the number of prediction masks
    \State $\tilde{y} \gets \{1 \text{ if } S_i = K, 0 \text{ otherwise} | i \in \text{all indices of } S\}$
    \State $M \gets \{1 \text{ if } S_i \neq 0 \text{ and } S_i \neq K, 0 \text{ otherwise} | i \in \text{all indices of } S\}$
    \State \Return $(\tilde{y}, M)$
\EndFunction
\end{algorithmic}
\end{algorithm*}

\begin{figure*}[h] 
   \centering 
   \includegraphics[width=1\textwidth]{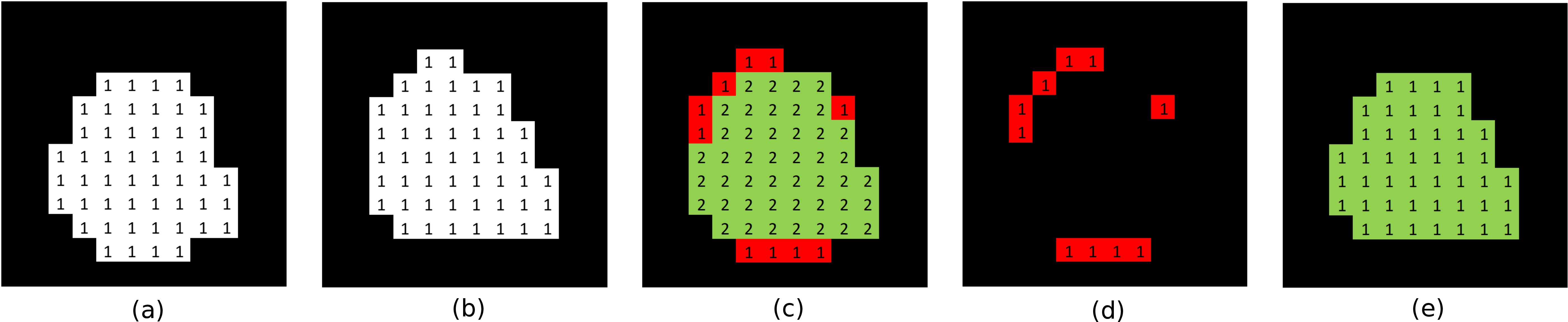} 
   \caption{Creation of a binary Inconsistency Mask (IM) from two teacher models. (a, b) Binary predictions from Models 1 and 2. (c) Sum of the prediction masks, highlighting areas of agreement (value 2) and disagreement (value 1). (d) The resulting IM marks only the regions of disagreement. (e) The final high-confidence pseudo-label contains only pixels where both models agreed.
} 
   \label{fig_IM_creation} 
\end{figure*}

For multi-class segmentation tasks, a similar logic applies, as detailed in Algorithm~\ref{alg_multiclass_im}.
\begin{itemize}
\item The final pseudo-label at each pixel is assigned the class label if and only if all $K$ models agree on that class. If there is any disagreement, the pixel is assigned a designated \texttt{ignore\_index} (or an explicit uncertainty class in Study B).
\item The Inconsistency Mask (IM) is a binary mask that marks all pixels where model predictions were not unanimous.
\end{itemize}

Multi-Label Logic: For multi-label tasks (e.g., HeLa) where outputs are independent sigmoids, we treat each class channel as an independent binary mask. A pixel is marked inconsistent in the final IM if any channel prediction differs across teachers.

\begin{algorithm*}
\caption{Multi-Class Predictions to Consensus Pseudo-Label and Inconsistency Mask}
\label{alg_multiclass_im}
\begin{algorithmic}[1]
\Require A non-empty list of hard prediction masks $pm = \{\hat{y}_1,\dots,\hat{y}_K\}$ with $K\ge2$ and uncertainty label $y_{\text{unc}}$.
\Function{PredMasksToImMulticlass}{$pm$, $y_{\text{unc}}$}
    \State $P \gets \text{stack}(pm)$
    \State Initialize $\tilde{y} \in \mathbb{Z}^{H \times W}$ and $M \in \{0,1\}^{H \times W}$
    \ForAll{pixels $p$}
        \If{$\forall k:\; P_{k,p} = P_{1,p}$}
            \State $\tilde{y}_p \gets P_{1,p}$
            \State $M_p \gets 0$
        \Else
            \State $\tilde{y}_p \gets y_{\text{unc}}$
            \State $M_p \gets 1$
        \EndIf
    \EndFor
    \State \Return $(\tilde{y}, M)$
\EndFunction
\end{algorithmic}
\end{algorithm*}

\subsection{Filtering Uncertainty with Inconsistency Masks}

The generated IM is the core of our filtering mechanism. For each unlabeled image, we create an input-pseudo-label training pair. The IM is applied to both the input image and the pseudo-label mask:
\begin{itemize}
\item Pixels in the input image corresponding to the IM are blacked out (set to zero). This strategy acts as a targeted form of CutOut~\cite{cutout} or Random Erasing~\cite{zhong_random_2017}, explicitly preventing the student model from learning features from visually ambiguous or difficult regions. 
\item Pixels in the pseudo-label mask corresponding to the IM are set to an \texttt{ignore\_index} for multi-class tasks or the background class for binary tasks. This explicitly prevents the student from being supervised by low-confidence targets.
\end{itemize}
This filtering process ensures the student model is trained on high-confidence, stable regions agreed upon by the entire teacher ensemble, directly combating error propagation.

\subsection{Morphological Refinements}

\begin{figure*}[h] 
   \centering 
   \includegraphics[width=1\textwidth]{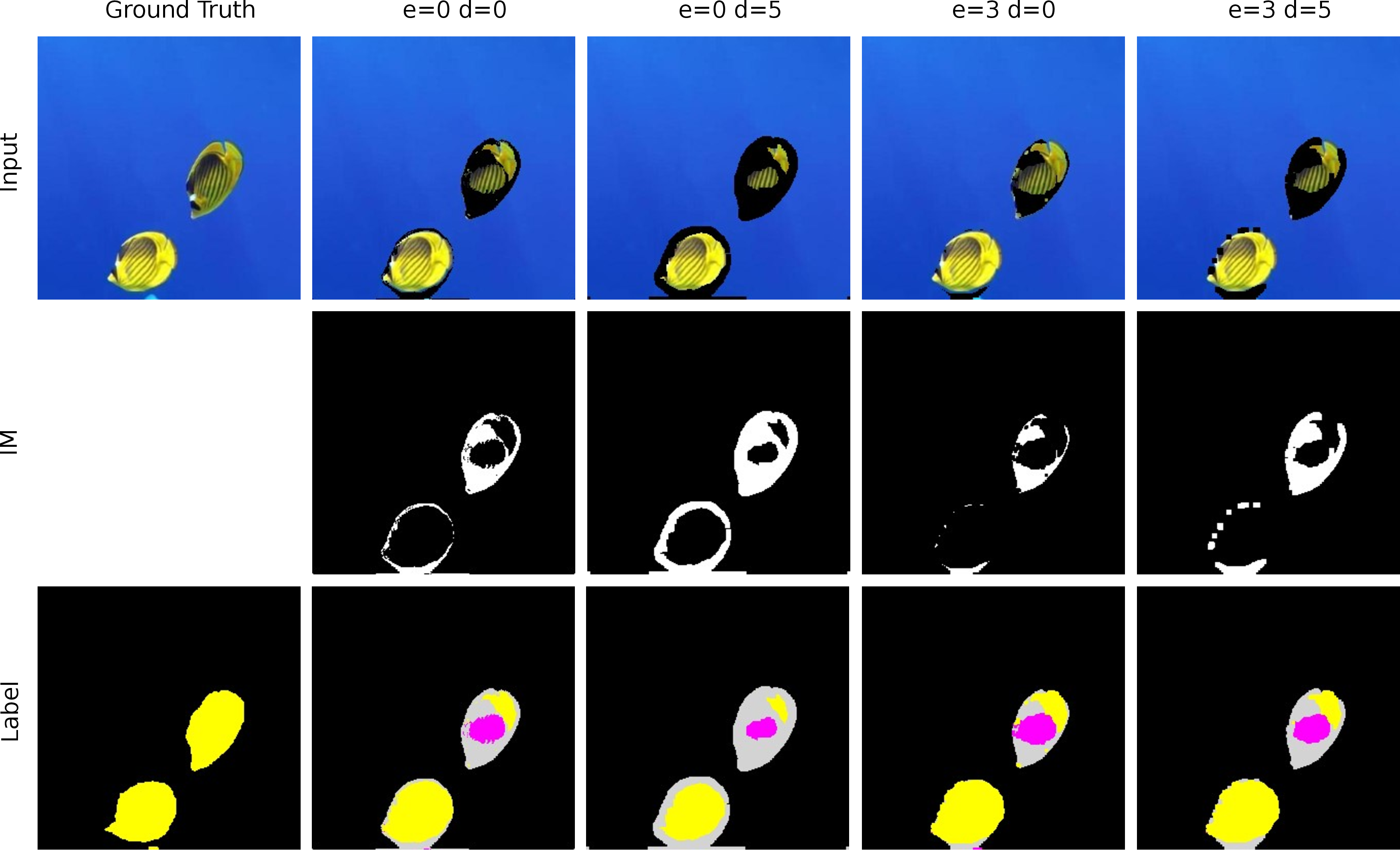} 
   \caption{Visualization of morphological operations on an image from the SUIM dataset. $e$ denotes erosion and $d$ dilation kernel sizes. A value of $0$ signifies that the operation is skipped. The first row displays the input image with the IM applied (masked regions set to black). The second row visualizes the IM structure under varying morphological parameters. The third row shows the corresponding pseudo-label masks: background (black), IM (gray), fish (yellow), and misclassified reef (magenta).} 
   \label{fig_morph_op} 
\end{figure*}

To fine-tune the spatial extent of the IM, we optionally apply morphological operations. An erosion operation can remove small, spurious regions of disagreement, while a dilation operation can expand the uncertain region to create a more conservative margin around ambiguous objects. As shown in our analysis, the optimal combination of these operations can be dataset-dependent. Figure~\ref{fig_morph_op} illustrates the effect of these operations.

\subsection{Generational SSL Framework}\label{subsubsec_im_gen}

We embed the IM mechanism within an iterative, generational self-training framework, which proceeds as follows:

\begin{enumerate}
  \item \textbf{Generation 0 (Bootstrap):} A set of initial models are trained exclusively on the small labeled dataset. The top-$K$ best-performing models on a validation set are selected as the initial "teacher ensemble".
  \item \textbf{Pseudo-Label Generation:} The teacher ensemble is used to predict labels for the entire unlabeled dataset. For each image, an IM and a corresponding high-confidence pseudo-label are generated as described in Section~\ref{subsubsec_im_def}.
  \item \textbf{Student Training:} A new set of student models is trained on the combined labeled and IM-filtered pseudo-labeled data. Each student's training is warm-started with the weights of the top model from the previous generation, promoting stable knowledge transfer.
  \item \textbf{Teacher Promotion:} At the end of training, the top-$K$ best-performing student models are selected to become the new teacher ensemble for the next generation.
  \item \textbf{Iteration:} Steps 2--4 are repeated for $G$ generations or until a convergence is reached.
\end{enumerate}

This discrete, generational process is a key element for stability. Unlike continuous EMA-based methods, it completely decouples the student training phase from the teacher's state, preventing the rapid feedback loops that can amplify noise.

\subsection{IM as a General SSL Performance Enhancer}\label{subsubsec_im_perf_enhanc}
To validate the broad applicability of our framework as a general performance enhancer, we apply Inconsistency Masks to existing state-of-the-art SSL methods (e.g., iMAS~\cite{iMAS}, U$^2$PL~\cite{u2pl}, UniMatch~\cite{UniMatch}). This setup strictly follows the Generational SSL Framework defined in Section~\ref{subsubsec_im_gen}, with a single modification to the Generation $0$ (Bootstrap) phase.

Instead of training the initial models exclusively on labeled data, we train them using the target SSL method (e.g., running the standard iMAS training loop). These SSL-trained models then serve as the teacher ensemble. From Generation $1$ onwards, the process reverts to the standard IM workflow: we generate Inconsistency Masks based on the teachers' disagreement and perform supervised training on the filtered pseudo-labels.

For the heterogeneous architecture of SegKC~\cite{segkc}, where the Senior (Large) and Junior (Small) models are co-trained in Generation $0$, we construct the teacher ensemble exclusively from the Junior model's checkpoints to generate the Inconsistency Masks and pseudo-labels. Furthermore, we strictly utilize the Junior model’s weights to initialize the student in Generation $1$. This ensures that the student architecture remains consistent throughout the generations and that performance gains are driven by the mask-based filtering rather than increased model capacity.

\section{Experiments}\label{Experiments}

We conducted two complementary sets of experiments to evaluate Inconsistency Masks (IM) across different regimes. First, we evaluate IM on the Cityscapes benchmark using standard pre-trained backbones to establish competitiveness with state-of-the-art methods and demonstrate IM's ability to act as a consistent performance booster. Second, we evaluate IM across four diverse datasets using a lightweight architecture trained entirely from scratch, testing the method's robustness in domains where pre-trained weights may not be applicable.

\subsection{Datasets and Task Definitions} 

Our evaluation spans four public datasets chosen for their diversity. ISIC 2018~\cite{tschandl_ham10000_2018, codella_skin_2019} provides a binary medical task. SUIM~\cite{islam_semantic_2020} offers a challenging underwater domain with 8 classes, characterized by severe visual ambiguity due to water turbidity, light scattering, and indistinct boundaries between organic textures (e.g., reefs fading into the seabed). HeLa (Ours) represents a highly specialized microscopy task. Finally, Cityscapes~\cite{cordts_cityscapes_2016} serves as the established standard benchmark for validating SSL performance. Crucially, we treat this dataset differently in our two studies to test different regime properties:

\begin{itemize}
    \item For Study A (Pre-trained Benchmarks), we use the standard 19-class subset to align with the established protocol of published SOTA results.
    \item For Study B (Training from Scratch), we use the full 34-class set (including void classes usually ignored) to evaluate performance in a high-class-count, fine-grained scenario.
\end{itemize}

For Study B, we utilize a 10\% labeled and 90\% unlabeled split for all datasets.

\begin{figure*}[h] 
   \centering 
   \includegraphics[width=1\textwidth]{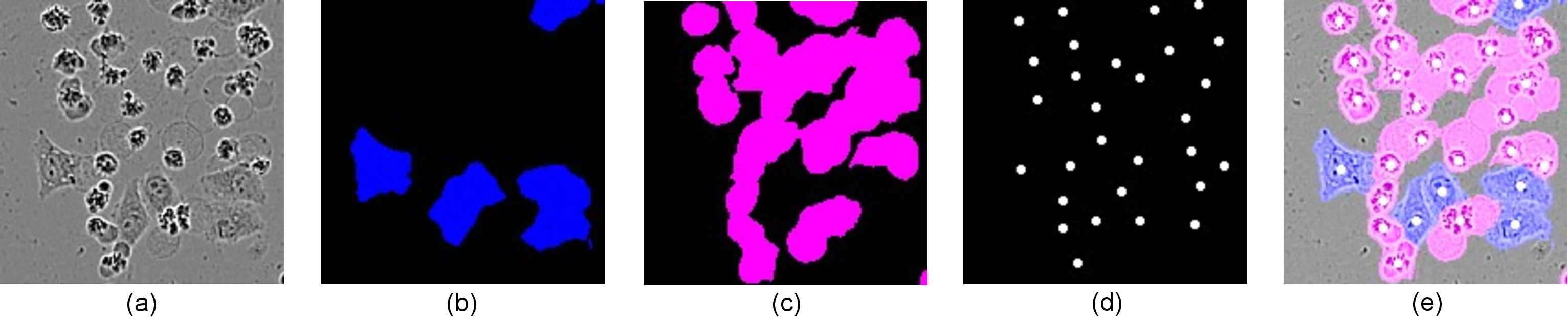} 
   \caption{The structure of the HeLa multi-label dataset. (a) The label-free brightfield image is the sole input to the model. The ground truth is decomposed into three independent masks: (b) ‘alive’ cells (blue), (c) ‘dead’ cells (magenta), and (d) ‘position’ points (white). (e) The final overlay demonstrates the multi-label nature of the task where masks overlap.} 
   \label{fig_hela_dataset} 
\end{figure*}

\subsection{The HeLa Multi-Label Dataset} 

As motivated in the Introduction, the HeLa dataset poses a non-standard multi-label segmentation challenge. Each sample consists of a $256 \times 256$ brightfield image and three corresponding binary masks for `alive' cells, `dead' cells, and cell `position'. Critically, these channels are not mutually exclusive; for example, a pixel can be both `alive' and `position'. This structure fundamentally deviates from standard multi-class segmentation assumptions and necessitates a model architecture with independent sigmoid outputs for each channel. This unconventional task structure, illustrated in Figure~\ref{fig_hela_dataset}, serves as a stringent test for the flexibility of the compared SSL methods.

To quantify performance on HeLa, we report Mean Cell Count Error (MCCE), defined as the average absolute per-image counting error summed over cell types (alive/dead), where lower is better (full definition in Appendix~\ref{sec_mcce_def}).

\subsection{Study A: Benchmarks with Pre-trained Backbones} 

To validate IM against state-of-the-art SSL methods, we utilized the Cityscapes benchmark. To ensure a strictly fair comparison, we reproduced all baseline methods within our unified codebase. We initialized models with backbones pre-trained on ImageNet (ResNet-50~\cite{he_deep_2016}) or, for foundational model comparisons, large-scale self-supervision (DINOv2~\cite{oquab_dinov2_2023}). We strictly adhered to the data partitions established by U$^2$PL~\cite{u2pl}, evaluating on $1/16$, $1/8$, $1/4$, and $1/2$ labeled splits.

For the teacher ensemble, we prioritized computational efficiency. Rather than training multiple independent models, we utilized a Checkpoint Ensemble strategy, selecting the top-$K$ best-performing checkpoints from a single training run. While we explored Snapshot Ensembles \cite{huang2017snapshotensemblestrain1} with cyclic learning rates, we observed that sufficient diversity for the Inconsistency Mask arises naturally during the standard training trajectory. Consequently, we found that complex restart schedules were unnecessary and achieved optimal results with a reduced budget of only $30$ epochs per generation.

\subsubsection{Enhancing State-of-the-Art Methods} 
Our primary goal was to determine if IM could improve the performance of modern SSL algorithms. We selected three leading methods that represent distinct paradigms in modern SSL: U$^2$PL~\cite{u2pl} (Contrastive Learning), iMAS~\cite{iMAS} (Adaptive Supervision), and UniMatch~\cite{UniMatch} (Dual-Stream Consistency). To bridge the gap between standard architectures and modern foundation models, we also extended our evaluation to UniMatch v2~\cite{UniMatch_v2}, which utilizes the massive DINOv2~\cite{oquab_dinov2_2023} Vision Transformer backbone.
To push the boundaries of our evaluation, we also applied IM to SegKC~\cite{segkc}, a recent method that employs knowledge distillation between a large ``Senior'' (DINOv2-Base) and a small ``Junior'' (DINOv2-Small) model.

\begin{table*}[t]
\centering
\caption{\textbf{Impact of Inconsistency Masks (IM) on Cityscapes validation performance.} We compare various baseline methods against their performance when augmented with our framework. Reported results for IM reflect the best performance achieved after generational convergence (typically Gen 2–3). All experiments use a Single-Scale + Sliding-Window inference protocol. Gains provided by IM are shown in parentheses. Note that all baseline results were re-implemented and trained in our environment to ensure a fair comparison.}
\label{tab_Study_A}
\begin{tabular}{@{}llcccc@{}}
\toprule
\textbf{Method} & \textbf{Backbone} & \textbf{1/16 Split} & \textbf{1/8 Split} & \textbf{1/4 Split} & \textbf{1/2 Split} \\
\midrule
\multicolumn{6}{l}{\textit{Standard Architectures (ResNet-50)}} \\
\midrule
Supervised Only & ResNet-50 & 64.93 & 70.20 & 74.22 & 77.65 \\
\textbf{+ IM (Ours)} & ResNet-50 & \textbf{72.53} \small{(+7.60)} & \textbf{74.47} \small{(+4.27)} & \textbf{77.95} \small{(+3.73)} & \textbf{78.78} \small{(+1.13)} \\
\midrule
U$^2$PL~\cite{u2pl} & ResNet-50 & 72.53 & 74.89 & 77.16 & 78.39 \\
\textbf{+ IM (Ours)} & ResNet-50 & \textbf{74.52} \small{(+1.99)} & \textbf{76.90} \small{(+2.01)} & \textbf{77.77} \small{(+0.61)} & \textbf{78.91} \small{(+0.52)} \\
\midrule
UniMatch~\cite{UniMatch} & ResNet-50 & 73.49 & 76.26 & 78.05 & 79.05 \\
\textbf{+ IM (Ours)} & ResNet-50 & \textbf{74.10} \small{(+0.61)} & \textbf{77.38} \small{(+1.12)} & \textbf{78.58} \small{(+0.53)} & \textbf{79.60} \small{(+0.55)} \\
\midrule
iMAS~\cite{iMAS} & ResNet-50 & 74.07 & 76.32 & 77.80 & 79.01 \\
\textbf{+ IM (Ours)} & ResNet-50 & \textbf{75.15} \small{(+1.08)} & \textbf{77.45} \small{(+1.13)} & \textbf{78.43} \small{(+0.63)} & \textbf{79.41} \small{(+0.40)} \\
\midrule
\multicolumn{6}{l}{\textit{Foundation Models (DINOv2)}} \\
\midrule
UniMatch v2~\cite{UniMatch_v2} & DINOv2-S & 80.67 & 81.71 & 82.32 & 82.84 \\
\textbf{+ IM (Ours)} & DINOv2-S & \textbf{80.97} \small{(+0.30)} & \textbf{81.93} \small{(+0.22)} & \textbf{82.59} \small{(+0.27)} & \textbf{83.07} \small{(+0.23)} \\
\midrule
SegKC~\cite{segkc} & DINOv2-S & 80.98 & 82.43 & 82.87 & 83.05 \\
\textbf{+ IM (Ours)} & DINOv2-S & \textbf{81.61} \small{(+0.63)} & \textbf{82.80} \small{(+0.37)} & \textbf{83.14} \small{(+0.27)} & \textbf{83.31} \small{(+0.26)} \\
\bottomrule
\end{tabular}
\end{table*}

As shown in Table~\ref{tab_Study_A}, the IM framework consistently improves the performance of every evaluated method across every data split. On the standard ResNet-50 backbone, IM boosts the strong iMAS and UniMatch baselines, pushing them to best-in-class performance levels. Even with the powerful DINOv2 backbone, which already achieves over 80\% mIoU, IM squeezes out further gains. It is important to contextualize these foundation model results: DINOv2-based baselines represent the current upper echelon of performance, where improvements are increasingly difficult to obtain due to the high quality of the foundational features. In this regime of diminishing returns, IM provides consistent improvements across all configurations, indicating that our uncertainty filtering captures a signal that is complementary even to massive-scale pre-training. Remarkably, IM further boosted the performance of the SegKC Junior model, demonstrating value even when the student is already benefiting from knowledge distillation from a larger Senior teacher. 

\begin{table*}[t]
\centering
\caption{\textbf{Training Efficiency on Cityscapes (1/16 Split).} We report the average time per epoch in seconds, including the rigorous sliding-window validation step. This split represents the maximum volume of unlabeled data ($\sim$94\% of dataset), creating the highest possible mask-generation overhead. Even when amortizing this offline cost over 30 epochs ($+70$s/epoch), IM is significantly faster than continuous SSL methods. Speedup values are calculated relative to the fastest competing method in each block (U$^2$PL for ResNet, UniMatch v2 for DINOv2).}
\label{tab_efficiency_study_a}
\small 
\setlength{\tabcolsep}{4pt} 
\begin{tabular}{lccc}
\toprule
\textbf{Method} & \textbf{Train} & \textbf{Total} & \textbf{Speedup} \\
 & \textbf{Time (s)} & \textbf{Time (s)} & \textbf{(vs. SOTA)} \\
\midrule
\multicolumn{4}{l}{\textit{Standard Architectures (ResNet-50)}} \\
U$^2$PL~\cite{u2pl} & 474 & 474 & 1.0$\times$ \\
iMAS~\cite{iMAS} & 483 & 483 & 0.98$\times$ \\
UniMatch~\cite{UniMatch} & 550 & 550 & 0.86$\times$ \\
\textbf{IM (Ours)} & 270 & \textbf{340} & \textbf{1.4$\times$} \\
\midrule
\multicolumn{4}{l}{\textit{Foundation Models (DINOv2-S)}} \\
UniMatch v2~\cite{UniMatch_v2} & 480 & 480 & 1.0$\times$ \\
SegKC~\cite{segkc} & 500 & 500 & 0.96$\times$ \\
\textbf{IM (Ours)} & 279 & \textbf{349} & \textbf{1.4$\times$} \\
\bottomrule
\end{tabular}
\end{table*}

\subsubsection{Training Efficiency Analysis} 

A potential concern for ensemble-based methods is the cost of training multiple teachers. However, our experiments show that for large, pre-trained models, full independent training is not required. By forming the teacher ensemble from the top-$K$ checkpoints of a single training run, we achieved the superior results in Table~\ref{tab_Study_A} with a computational budget comparable to standard single-model training.

Beyond this setup efficiency, Table~\ref{tab_efficiency_study_a} demonstrates that IM is also significantly faster during training than current state-of-the-art SSL pipelines. Continuous methods like UniMatch and iMAS incur a high online overhead because they require forward passes for both the student and the teacher (or EMA model) at every iteration. In contrast, IM moves the teacher inference to an offline "Generation" phase. We evaluated this efficiency on the Cityscapes 1/16 split, which contains the largest amount of unlabeled data (94\% of the dataset) and thus represents the maximum computational overhead for our method. Even when amortizing this worst-case mask generation time (approx. 35 minutes, or 70 seconds/epoch) into the training budget, IM reduces the effective training time by 30--40\% compared to UniMatch and iMAS. By decoupling pseudo-label generation from training, IM effectively converts the SSL problem into a standard supervised training loop, allowing for significantly higher throughput while maintaining superior accuracy.

\subsection{Study B: Resource-Constrained Regimes} \label{sec_study_b_results}

In many specialized domains, pre-trained backbones may be suboptimal due to large domain shifts or computationally too heavy for deployment. To evaluate IM in this challenging ``from scratch'' regime, we utilized our lightweight $1\times1$ U-Net (detailed in Appendix~\ref{Appendix_A}) initialized with random weights. Given the high variance of training small models from scratch, we formed teacher ensembles by selecting the best models from independently trained runs to ensure maximum diversity.

We compared IM against seven SSL methods: FixMatch~\cite{sohn2020fixmatch}, Fuzzy Positive Learning (FPL)~\cite{fuzzypositivlearning}, CrossMatch~\cite{zhao2024crossmatch}, iMAS~\cite{iMAS}, UPS~\cite{UPS}, UniMatch~\cite{UniMatch} and U$^2$PL~\cite{u2pl}. To contextualize their performance, we established four reference points:

\begin{itemize}
    \item \textbf{Labeled Data Training (LDT):} A baseline model trained only on the 10\% labeled data. All SSL methods were warm-started from the best LDT model to ensure a fair initial condition. We selected this unaugmented baseline as the starting point to clearly distinguish the performance boost provided by the SSL algorithms from the gains achievable through data augmentation alone (represented by ALDT)
    \item \textbf{Augmented LDT (ALDT):} A stronger baseline trained on a strongly  augmented version of the labeled data.
    \item \textbf{Full Dataset Training (FDT):} A fully supervised reference baseline trained on the complete 100\% labeled dataset, included to contextualize the performance gap between our low-data regime and a full-data scenario.
    \item \textbf{Augmented FDT (AFDT):} A fully supervised baseline trained on the complete 100\% labeled dataset with strong augmentations, serving as strong fully-supervised reference point.
\end{itemize}
    
Note that several methods required bespoke adaptations to handle the multi-label HeLa dataset (detailed in Appendix~\ref{Appendix_A_Adaptions}).

\subsubsection{Methodological Robustness across Diverse Domains} 
\label{sec_analysis_robustness}
Table~\ref{tab_Study_B} summarizes the peak performance achieved for each method. For our iterative IM framework, we report the best result across five generations. For continuous baselines (e.g., FixMatch, iMAS), which do not use generations, we report the best validation score achieved during a fixed 50-epoch training run.

\begin{table}[b]
\caption{Best Achieved Performance (mIoU / MCCE) in the Generalization Study. 
(For HeLa, lower MCCE is better. Best SSL results are bolded. UPS failed to run on Cityscapes due to high memory requirements.)}
\label{tab_Study_B}
\footnotesize 
\setlength{\tabcolsep}{4pt} 

\begin{tabular}{@{}lllll@{}}
\toprule
  & \makecell[l]{ISIC 2018 \\ (IoU~$\uparrow$)} 
  & \makecell[l]{HeLa \\ (MCCE~$\downarrow$)} 
  & \makecell[l]{SUIM \\ (mIoU~$\uparrow$)} 
  & \makecell[l]{Cityscapes \\ (mIoU~$\uparrow$)} \\
\midrule
LDT         & 67.1    & 9.9     & 35.7   & 32.0 \\
ALDT        & 72.4    & 3.3     & 43.2   & 37.4 \\
FDT         & 75.1    & 2.5     & 51.7   & 45.6 \\
AFDT        & 77.3     & 2.4     & 52.7    & 45.8 \\
\addlinespace
FixMatch\cite{sohn2020fixmatch}         & 70.3    & 42.6    & 36.1   & 36.6 \\
FPL\cite{fuzzypositivlearning}          & 68.4    & 30.6    & 25.7   & 15.2 \\
CrossMatch\cite{zhao2024crossmatch}     & 65.7    & 3.6     & 36.5   & 34.7 \\
iMAS\cite{iMAS}                         & 66.1    & 13.8    & 33.7   & 35.2 \\
UPS\cite{UPS}                           & 37.9    & 24.9    & 26.8   & -    \\
U$^2$PL\cite{u2pl}                      & 67.5    & 22.6    & 36.6   & 35.5 \\
UniMatch\cite{UniMatch}                 & 64.0    & 7.7     & 26.5   & 24.3 \\
\addlinespace
ME          & 69.0    & 3.9     & 37.1   & 35.0   \\
IM          & \textbf{72.3}    & \textbf{2.8}     & \textbf{44.3}   & \textbf{40.7} \\
\bottomrule
\end{tabular}
\end{table}

In this resource-constrained setting, IM consistently outperforms all other SSL methods. We observed distinct failure modes among the competing algorithms. Crucially, despite being warm-started with the converged weights of the LDT model, several methods struggled to surpass--or even maintain--this supervised baseline. While iMAS remained stable, it frequently stagnated, failing to extract significant gains from the unlabeled data. Conversely, confidence-based methods like FixMatch and FPL struggled with error propagation on ambiguous datasets like SUIM, exhibiting severe performance collapse after initial gains. Notably, IM is the only SSL method to consistently match (within 0.1 mIoU) or surpass the stronger ALDT baseline across all four datasets. 
Remarkably, on ISIC 2018, the advanced IM++ variants (detailed in Appendix~\ref{Appendix_B_IM_Advanced_Variants}) surpass the standard Full Dataset Training (FDT) baseline. However, when we apply heavy augmentations to the full dataset (AFDT), the performance ceiling rises further, confirming that while IM extracts maximum value from limited data, the combination of full supervision and strong augmentation remains the absolute upper bound.

Finally, we compared our approach against a standard Model Ensemble (ME), which averages the predictions of the teacher models rather than filtering them. While ME improves over the single-model baseline (e.g., 69.0\% vs 67.1\% on ISIC 2018), it consistently underperforms compared to IM (72.3\%). This gap is even more pronounced on the challenging SUIM dataset (37.1\% vs 44.3\%). This confirms that filtering based on disagreement is substantially more effective than averaging out disagreement, as standard ensembling retains high-confidence errors while IM explicitly removes them.

\begin{figure*}[htbp] 
   \centering 
   \includegraphics[width=1\textwidth]{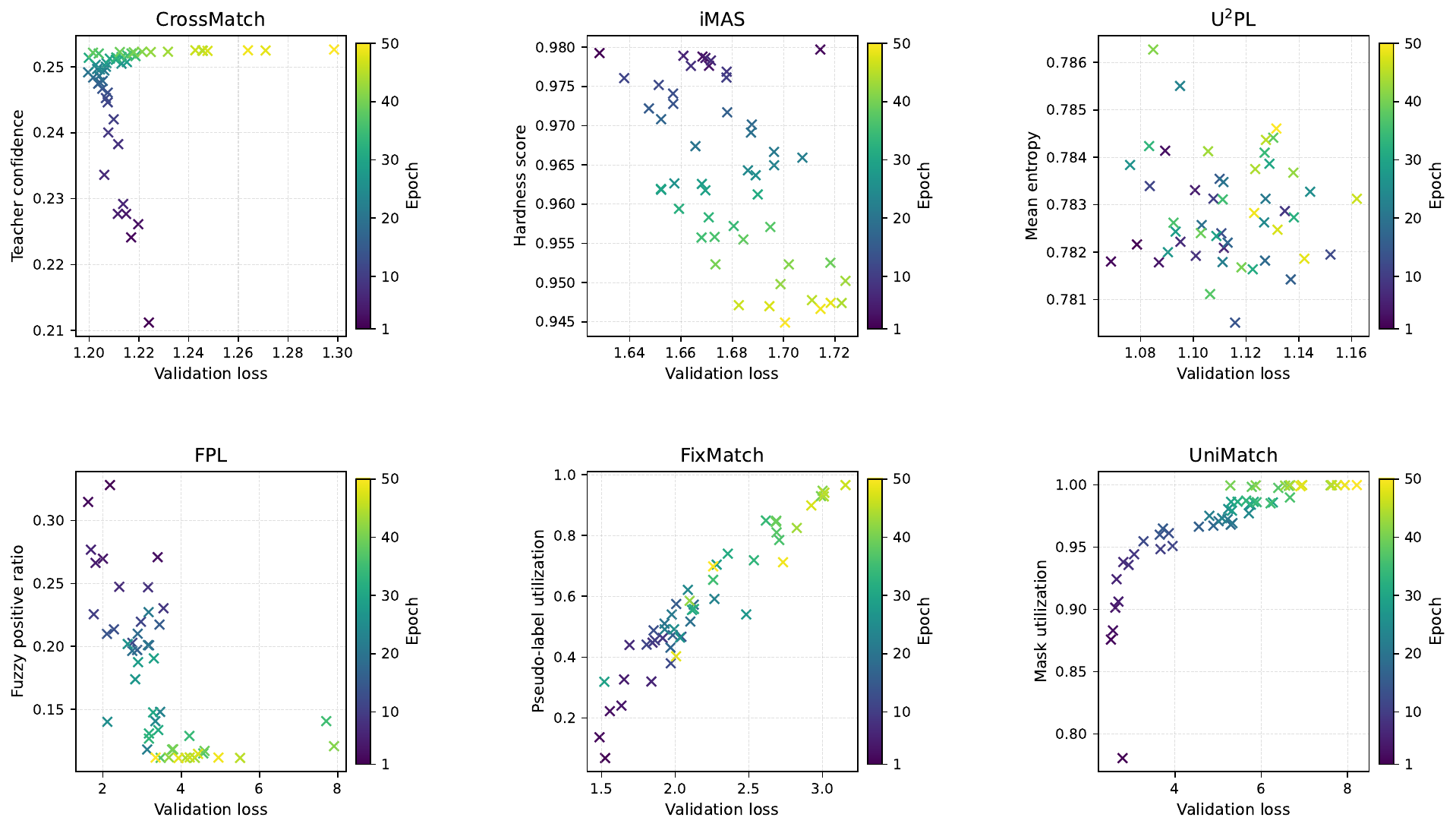} 
   \caption{Analysis of training dynamics on the SUIM dataset. We plot validation loss (x-axis, lower is better) against method-specific internal metrics (y-axis) over 50 epochs. Color indicates training progress from early (purple) to late (yellow). The trajectories visualize the stability profiles discussed in Sec.~\ref{sec_dynamics_analysis}: FixMatch and FPL exhibit pathological divergence (where internal training dynamics decouple from validation performance), U$^2$PL shows noisy guidance, CrossMatch improves systematically before saturating, while iMAS remains stable but largely stagnant.} 
   \label{fig_instab} 
\end{figure*}

\subsubsection{Analysis of Training Dynamics and Stability}\label{sec_dynamics_analysis} 

To understand why certain methods succeed or fail in this resource-constrained regime, we analyzed their internal training dynamics. Continuous methods (FixMatch, FPL, iMAS, U$^2$PL, CrossMatch) rely on internal metrics---such as pseudo-label confidence or entropy---to guide learning. Ideally, an improvement in these metrics should correlate with an improvement in validation performance.

Figure~\ref{fig_instab} plots the validation loss against these core metrics over 50 epochs on the SUIM dataset. Points are colored by epoch (dark=early, yellow=late). A critical observation is that for several methods, the optimal performance (lowest validation loss) is achieved in the early epochs (dark points), after which the SSL mechanism drives the model away from this optimum.
The trajectories reveal three distinct stability profiles:

\begin{itemize}
    \item \textbf{Pathological Dynamics (FixMatch, FPL \& UniMatch):} These methods suffer from severe confirmation bias and model collapse.
    \begin{itemize}
        \item \textbf{FixMatch:} As training progresses, the model becomes more confident (utilization increases), but validation loss worsens. The model learns to be confidently wrong on ambiguous underwater features.
        \item \textbf{UniMatch:} Mirrors the failure mode of FixMatch but often more severely. Without a pre-trained backbone to ground the features, the dual-stream consistency mechanism reinforces hallucinations, leading to performance significantly below the supervised baseline on complex datasets like SUIM.
        \item \textbf{FPL:} Shows a catastrophic divergence where the loss skyrockets in later epochs, indicating the fuzzy-positive mechanism fails to provide a coherent signal in this setting.
    \end{itemize}
    
    \item \textbf{Noisy Guidance (U$^2$PL):} The trajectory appears as a scattered cloud. The teacher’s entropy fluctuates without a clear correlation to student performance, indicating the signal is too noisy to guide the model consistently.
    
    \item \textbf{Stable but Stagnant (iMAS \& CrossMatch):} These methods show non-divergent dynamics but limited capacity for improvement.
    \begin{itemize}
        \item \textbf{CrossMatch:} Exhibits a healthy initial correlation where rising teacher confidence corresponds to lower validation loss. However, it quickly saturates, reaching a point where confidence continues to rise while performance plateaus.
        \item \textbf{iMAS:} Shows a very gradual linear trajectory. While stable, the validation loss changes minimally throughout training, indicating that the adaptive supervision mechanism struggles to extract significant new information from the unlabeled data in this regime.
    \end{itemize}
\end{itemize}

These dynamics directly explain the performance discrepancy seen in Table~\ref{tab_Study_B}. For instance, FixMatch performs exceptionally well on Cityscapes-34 and ISIC (tasks with clear boundaries), where high confidence correlates with accuracy. However, on SUIM, where underwater boundaries are ambiguous, high confidence is often misplaced. The trajectory plot confirms that FixMatch diverges over time; it is a high-risk strategy that lacks an internal brake.

In contrast, Inconsistency Masks succeed because the generational framework inherently implements a ``stop-and-reevaluate'' mechanism. By filtering based on disagreement (a spatial consensus metric) rather than confidence (a single-model metric), IM prevents the divergence seen in FixMatch. This distinction is crucial in the absence of pre-trained weights: while methods relying on confidence or entropy implicitly depend on the calibrated features of strong backbones to guide early training, IM effectively filters noise even when feature representations are nascent, preventing the model collapse observed in continuous baselines.

\section{Conclusion}

In this work, we addressed a fundamental challenge in semi-supervised learning: the training instability and error propagation caused by noisy pseudo-labels. We introduced Inconsistency Masks (IM), a novel framework that reframes model disagreement not as noise to be averaged away, but as a powerful, explicit signal for identifying and filtering uncertainty. By removing ambiguous regions from the training process, this simple mechanism effectively decouples the student from the teacher's errors, breaking the cycle of confirmation bias that plagues many SSL methods.

On the competitive Cityscapes benchmark (Study A), IM acted as a general enhancement framework: when applied to leading approaches like iMAS, U2PL, and UniMatch, it consistently boosted accuracy by filtering residual noise from their predictions, achieving superior benchmarks across ResNet-50 and DINOv2-based architectures. Furthermore, our resource-constrained study (Study B) confirmed the method's robustness on diverse datasets from medical and underwater domains. Even when trained entirely from scratch without pre-trained backbones, IM outperforms other SSL methods, avoiding the divergence and stagnation observed in competing approaches.

In contrast to the erratic behavior of continuous training loops, our decoupled, generational framework ensures consistent  improvement. By prioritizing direct uncertainty filtering, IM offers a reliable solution for specialized domains while simultaneously serving as a plug-and-play enhancement for state-of-the-art foundation models.

We believe harnessing disagreement opens a promising avenue for dependable semi-supervised learning beyond 2D segmentation. Future work could extend IM to regression tasks like image or audio denoising by defining masks via prediction variance (e.g., a $2\sigma$ threshold), or apply it to complex structures such as 3D voxels.

\section{Acknowledgments}

The first author (M.R.H.V.) would like to extend his heartfelt gratitude to the Deep Learning and Open Source Community, particularly to Dr. Sreenivas Bhattiprolu, Harrison Kinsley, and the team at Deeplizard (Chris and Mandy). Their high-quality educational resources and shared wisdom were instrumental in his self-education in computer science and deep learning. This work would not have been possible without these open and free resources. 

Large Language Models (Gemini and ChatGPT) were used for translation, editing, and grammatical refinement of the manuscript.

\section{Declarations}

\begin{itemize}
    \item Funding: The creation of the HeLa Dataset was funded by the Deutsche Forschungsgesellschaft (DFG, German Research foundation), SFB1403 – project no. 414786233 and SPP2306 – project no. GA1641/7-1 and by the Bundesministerium für Bildung und Forschung (BMBF) project. 16LW0213.
    \item Conflict of interest/Competing interests: The authors have no other relevant financial or non-financial interests to disclose.
    \item Consent for publication: All authors consent to this publication.
    \item Code and Data Availability: The full source code, along with the newly created HeLa dataset, is available at: \url{https://github.com/MichaelVorndran/InconsistencyMasks}. The public benchmark datasets analyzed in this study are available from their respective official sources: Cityscapes~\cite{cordts_cityscapes_2016}, ISIC 2018~\cite{codella_skin_2019}, and SUIM~\cite{islam_semantic_2020}.
\end{itemize}

\clearpage
\onecolumn 

\newpage

\begin{appendices}
\counterwithin{figure}{section}
\counterwithin{table}{section}

\section{Architectures and Implementation Details} \label{Appendix_A}
\subsection{The $1\times1$ U-Net Architecture} 
\begin{figure}[b]
    \centering
    \begin{minipage}{0.48\textwidth}
        \centering
        \includegraphics[width=\textwidth]{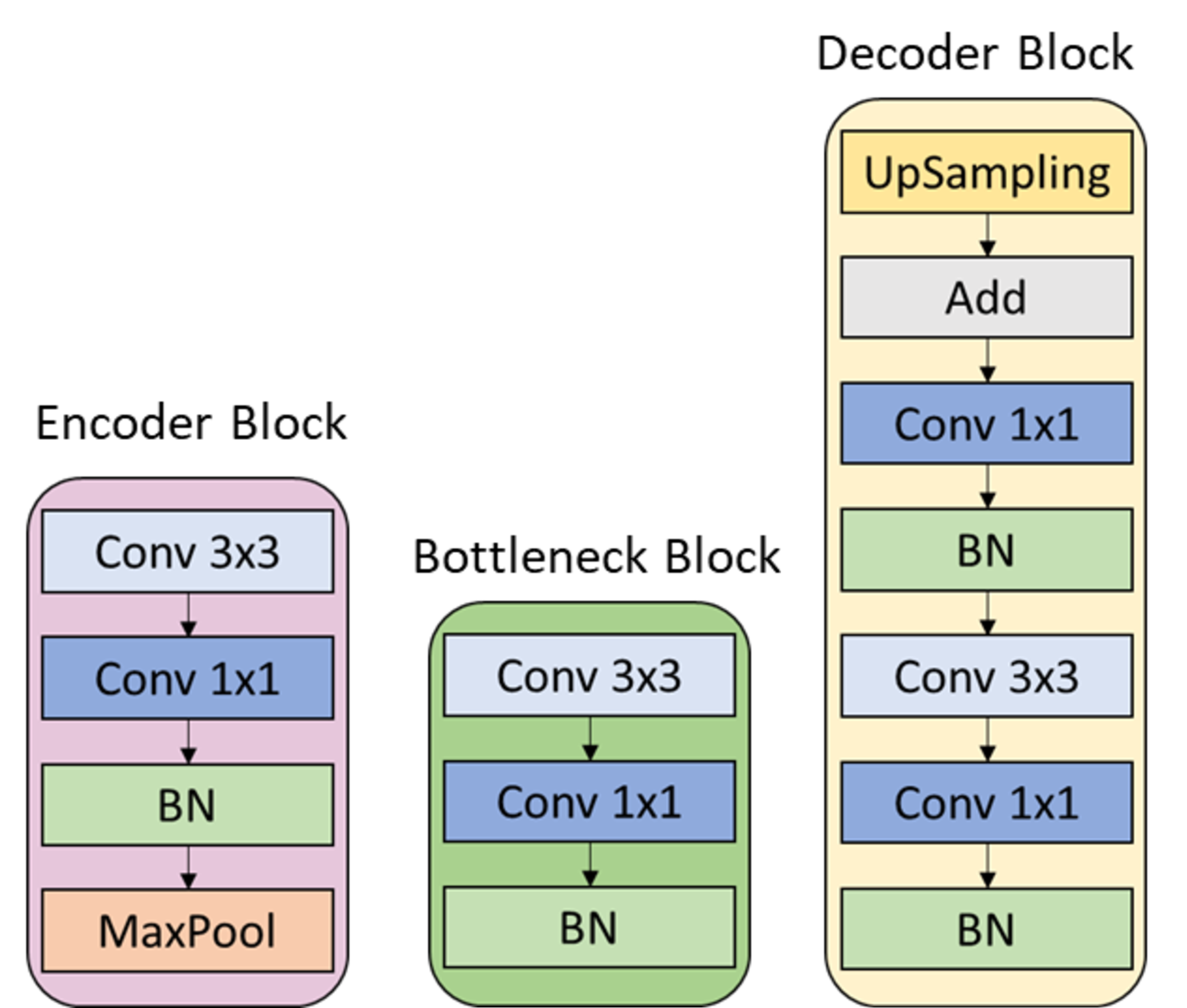}
        \caption{Building blocks of the $1\times1$ U-Net.}
        \label{fig_1x1_building_blocks}
    \end{minipage}
    \hfill
    \begin{minipage}{0.48\textwidth}
        \centering
        \includegraphics[width=\textwidth]{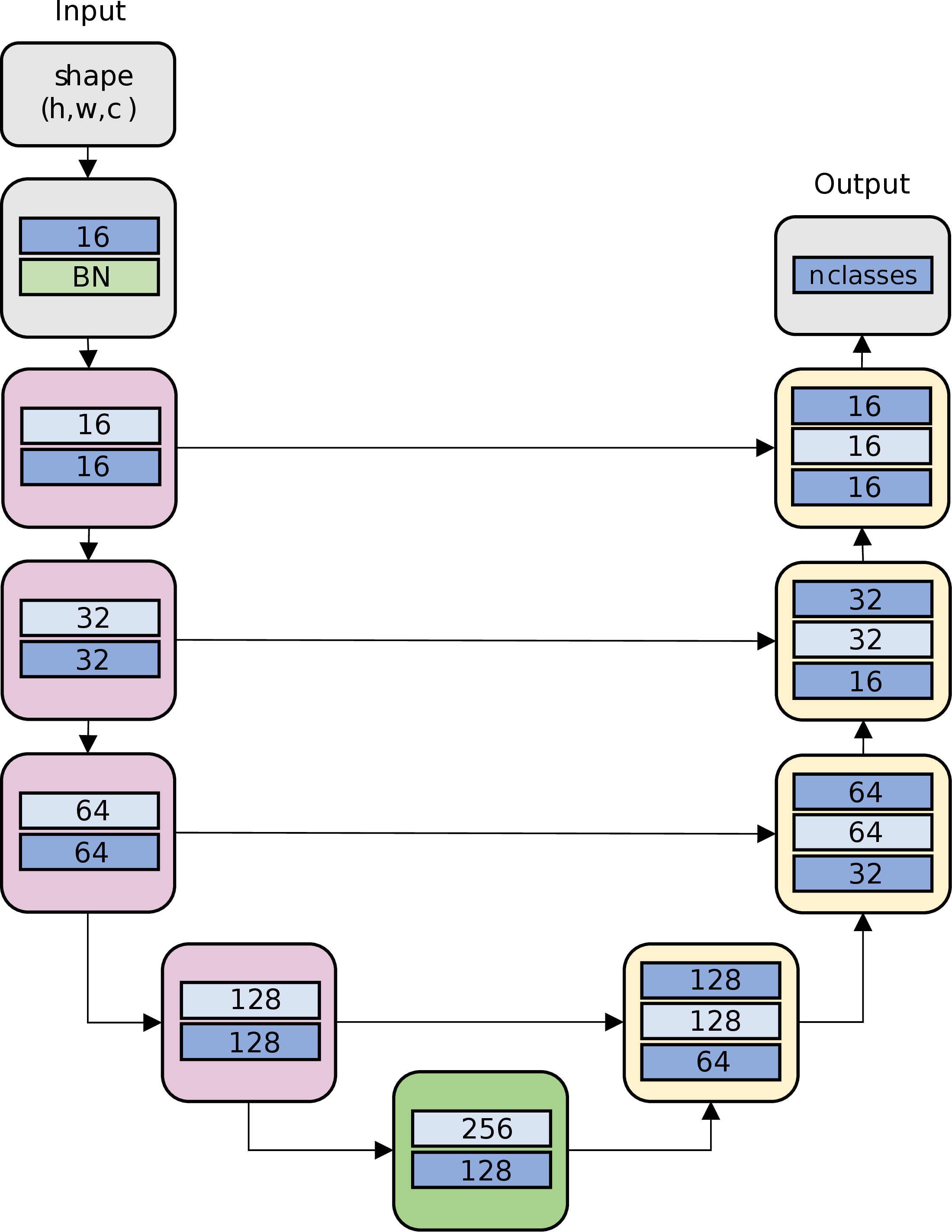}
        \caption{Conceptual depiction of our $1\times1$ U-Net architecture.}
        \label{fig_1x1_arch}
    \end{minipage}
\end{figure}

For our Generalization \& Stability Study, we designed a lightweight U-Net variant, termed ``$1\times1$ U-Net'', optimized for fast training from scratch under resource-constrained conditions. Our key modification from the standard U-Net~\cite{ronneberger_u-net_2015} is the inclusion of a $1\times1$ convolutional layer following every $3\times3$ convolution, a technique inspired by the Inception network~\cite{szegedy_going_2014}. This allows for dimensionality reduction and feature pooling across channels, enhancing representational capacity without increasing the parameter count. The model's width and complexity can be adjusted via a scaling factor $\alpha$, which controls the number of filters in each layer. A detailed diagram of the architecture's building blocks is shown in Figure~\ref{fig_1x1_building_blocks}, and the full architecture is depicted in Figure~\ref{fig_1x1_arch}, with its complexity profile provided in Table~\ref{tab_unet_complexity}.

\begin{table}[h]
\centering
\caption{Comparison of $1\times1$ U-Net complexity (varying width factor $\alpha$) versus recent reference models in terms of Parameters (Millions) and FLOPs (Billions).}
\label{tab_unet_complexity}
\setlength{\tabcolsep}{8pt} 
\begin{tabular}{@{}lcc@{}}
\toprule
\textbf{Model / Configuration} & \textbf{Params (M)} & \textbf{FLOPs (G)}\\
\midrule
\textit{$1\times1$ U-Net (Ours) - Scaling $\alpha$} & & \\
$\alpha = 0.5$ & 0.17 & 0.006 \\
$\alpha = 0.75$ & 0.38 & 0.014 \\
$\alpha = 1.0$ & 0.68 & 0.025 \\
$\alpha = 1.25$ & 1.06 & 0.039 \\
$\alpha = 1.5$ & 1.53 & 0.056 \\
$\alpha = 1.75$ & 2.08 & 0.076 \\
$\alpha = 2.0$ & 2.72 & 0.099 \\
\midrule
\textit{Reference Models} & & \\
Attention U-Net~\cite{oktay_attention_2018} & 7.73 & - \\
ConvNeXt-S~\cite{liu_convnet_2022} & 22.00 & 4.3 \\
ViT-S~\cite{dosovitskiy_image_2020} & 22.00 & 4.6 \\
U-Net (Original)~\cite{ronneberger_u-net_2015} & 31.00 & 16.6 \\
DeepLabV3+ (Xception) ~\cite{deeplabv3+} & 41.00 & 40.0 \\
\bottomrule
\end{tabular}
\end{table}

\subsubsection{Input Block}
The input block starts with the normalization of pixel intensities. This is followed by a $1\times1$ convolutional layer and then batch normalization.

\subsubsection{Encoder}
The encoder comprises four blocks, each containing a $3\times3$ convolutional layer followed by a $1\times1$ convolutional layer. The number of filters in these layers doubles with each subsequent block. Each block concludes with batch normalization and max pooling.

\subsubsection{Bottleneck}
The bottleneck contains a $3\times3$ convolutional layer followed by a $1\times1$ convolutional layer. Batch normalization concludes the block.

\subsubsection{Decoder}
The structure of the decoder mirrors the encoder, consisting of four blocks. Each block initiates with the upsampling of the feature map from the preceding deeper layer. The resulting feature map is then merged with the feature map from the corresponding encoder block through an addition operation. Subsequently, a $1\times1$ convolutional layer is applied, followed by batch normalization. This sequence is succeeded by a $3\times3$ convolutional layer, another $1\times1$ convolutional layer, and a final round of batch normalization to conclude the block.

\subsubsection{Output Block}
The output block applies a $1\times1$ convolutional layer to the output from the final decoder block. The choice of activation function for this layer, as well as the number of output classes, is specified based on the task at hand.

\subsection{The EvalNet Architecture} ~\label{Appendix_A_EvalNet}
\begin{figure}[h] 
   \centering 
   \includegraphics[width=0.2\textwidth]{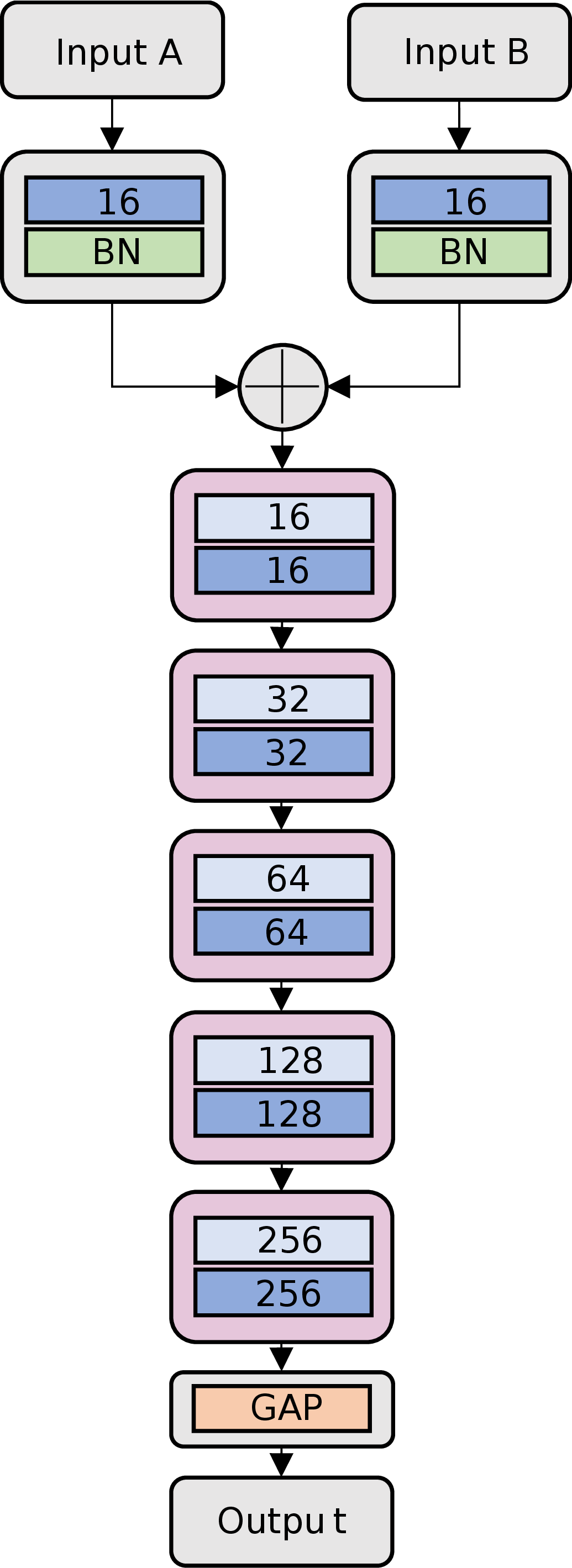} 
   \caption{Conceptual depiction of the EvalNet architecture} 
   \label{fig_EvalNet} 
\end{figure}

The EvalNet, used in our IM++ and EvalNet-based experiments, is a regression model designed to predict the Intersection over Union (IoU) score of a given input image and its predicted segmentation mask. As illustrated in Figure~\ref{fig_EvalNet}, it consists of two parallel encoder streams that process the image and mask independently. Their features are then concatenated and passed through a series of convolutional blocks. The final output is a scalar IoU prediction produced by a global average pooling layer followed by a dense layer. For multi-class tasks, the head is modified to predict a per-class IoU and a class presence probability, enabling a more nuanced quality assessment.

\subsection{Adaptations for Multi-Label Segmentation}\label{Appendix_A_Adaptions}

The HeLa dataset's multi-label nature, where each pixel is independently classified across three channels (`alive', `dead', `position'), required non-trivial modifications to several SSL methods that assume a single-label, multi-class output space. Our adaptations, detailed below, were designed to preserve the core logic of each method while accommodating the sigmoid-based, per-channel predictions.

\begin{itemize}
    \item \textbf{FixMatch~\cite{sohn2020fixmatch} \& CrossMatch~\cite{zhao2024crossmatch}:} These methods were adapted by treating the task as three parallel binary segmentation problems. For FixMatch, the confidence threshold $\tau$ was applied independently to each channel's sigmoid output. A pseudo-label for a given channel was generated if its probability was above $\tau$ (positive) or below $1-\tau$ (negative), allowing each channel to be supervised independently. For CrossMatch, the distillation losses were calculated using a channel-wise Dice loss between the teacher and student sigmoid probabilities.
    
    \item \textbf{U$^2$PL~\cite{u2pl}:} The core uncertainty signal in U$^2$PL is entropy. A standard multi-class entropy assumes mutually exclusive classes. To correctly adapt this to our multi-label task, we modeled the three output channels as independent Bernoulli distributions. The total uncertainty (joint entropy) for a pixel is therefore the sum of the individual Bernoulli entropies of each channel. This approach correctly measures the total ambiguity across the three independent classification decisions for each pixel. For the contrastive learning component, the binary on/off state of the three channels was encoded into a unique ``combination ID'' (an integer from $0$ to $7$), and contrastive learning was performed between pixel representations and prototypes for these eight possible label combinations.
    
    \item \textbf{iMAS~\cite{iMAS}:} The hardness metric in iMAS is derived from the agreement between student and teacher. We re-formulated this for the multi-label case by first binarizing the sigmoid outputs of both models at $0.5$. We then calculated a symmetric, class-frequency-weighted average of the per-channel Intersection over Union (IoU). This effectively measures agreement across all three independent segmentation tasks to produce a single hardness score for each instance.
    
    \item \textbf{FPL (Fuzzy Positive Learning)~\cite{fuzzypositivlearning}:} FPL is fundamentally designed for multi-class softmax outputs. We adapted its core idea by creating ``fuzzy target values'' for a Mean Squared Error (MSE) loss. For pixels where the teacher was confident (probability $> \tau$ or $< 1-\tau$), a hard $0$ or $1$ was used as the target. For uncertain pixels, the teacher's soft sigmoid probability itself (e.g., $0.63$) was used as a ``fuzzy'' target, encouraging the student to mimic the teacher's uncertainty distribution.

    \item \textbf{UniMatch~\cite{UniMatch}:} The core dual-stream and feature perturbation topology was preserved. We adapted the loss to a channel-wise Binary Cross-Entropy (BCE). The confidence mask was generated independently for each channel: a pixel was considered valid if the teacher's probability was $> \tau$ (positive) or $< 1-\tau$ (negative). Consistent with the original method, we used hard pseudo-labels (thresholded at 0.5) for the student's supervision targets.
\end{itemize}

These necessary and bespoke adaptations underscore a key advantage of our IM framework: its inherent dataset-agnosticism. IM's logic of comparing discrete predictions for disagreement operates identically on binary, multi-class, or multi-label outputs without requiring changes to its core algorithm.

\subsection{Hyperparameters and Training Protocols}
To ensure reproducibility and fair comparison, we provide detailed training protocols for both experimental regimes. Study A focuses on high-performance benchmarking using pre-trained backbones on Cityscapes, while Study B targets resource-constrained scenarios with lightweight architectures trained from scratch across diverse domains.

\subsubsection{Study A Protocols}
Table~\ref{tab_hyperparams_Study_A} details the hyperparameter settings for the Cityscapes benchmarks. In \textbf{Gen 0 (Baseline SSL)}, we train the respective method (e.g., iMAS, U$^2$PL, UniMatch) using its original training loop and loss functions. In \textbf{Gen 1+ (IM Self-Training)}, we switch to our unified supervised training loop using Inconsistency Masks.

Regarding input resolution and backbone configuration, we enforced consistency across all ResNet-50 based methods. We utilize a standard output stride of 16 to balance computational efficiency with spatial resolution. While the original UniMatch implementation utilizes a crop size of $801 \times 801$, we standardized this to $768 \times 768$ to match the settings of iMAS and U$^2$PL, ensuring that performance differences are attributable to the method rather than resolution advantages. For DINOv2-based methods (UniMatch v2, SegKC), we used $798 \times 798$ to align with the ViT patch structure.

\begin{table}[h]
\centering
\caption{Training Hyperparameters for Study A (Cityscapes SOTA)}
\label{tab_hyperparams_Study_A}
\begin{tabular}{lll}
\toprule
\textbf{Parameter} & \textbf{Gen 0 (Baseline SSL)} & \textbf{Gen 1+ (IM Self-Training)} \\ 
\midrule
Primary Optimizer & AdamW & AdamW \\
Backbone LR & $5 \times 10^{-6}$ (DINO) / $1 \times 10^{-2}$ (ResNet) & $0.1 \times \text{Initial LR}$ \\
Decoder/Head LR & $2 \times 10^{-4}$ (DINO) / $1 \times 10^{-2}$ (ResNet) & $0.1 \times \text{Initial LR}$ \\
Batch Size & 6 -- 16 (Method Dependent) & 16 \\
Crop Size (ResNet) & 768 $\times$ 768 & 768 $\times$ 768 \\
Crop Size (DINOv2) & 798 $\times$ 798 & 798 $\times$ 798 \\
Training Epochs & 150 -- 200 & 30 \\
EMA Alpha & 0.996 -- 0.999 & 0.99 \\
Dropout Rate & 0.0 & 0 -- 0.3 \\
Loss Function & OHEM~\cite{ohem} / Method-Specific & CE + Lovász Softmax~\cite{berman2018lovasz} \\
\midrule
\multicolumn{3}{l}{\textit{Inconsistency Mask Parameters}} \\
Number of Teachers & - & 2 \\
IM Erosion Kernel & - & $3 \times 3$ \\
IM Dilation Kernel & - & $5 \times 5$ \\
\midrule
Hardware & \multicolumn{2}{l}{Google Colab A100 (40GB) Instance} \\ 
\bottomrule
\end{tabular}%
\end{table}

\subsubsection{Study B Protocols}

Table~\ref{tab_hyperparams_Study_B} outlines the configuration for the resource-constrained experiments. Here, models are trained from scratch using our $1\times1$ U-Net architecture. We employ a fixed budget of 50 epochs per generation to monitor stability and prevent overfitting in the absence of large pre-training datasets.

\begin{table}[h]
\centering
\caption{Training Hyperparameters for Study B (Resource-Constrained)}
\label{tab_hyperparams_Study_B}
\begin{tabular}{ll}
\toprule
\textbf{Parameter} & \textbf{Configuration Value} \\ 
\midrule
\textbf{Input Resolutions ($H \times W$)} & \\
\hspace{1em} ISIC 2018 / HeLa / SUIM & $256 \times 256$  \\
\hspace{1em} Cityscapes & $208 \times 416$  \\
\midrule
Architecture & $1\times1$ U-Net (see Appendix~\ref{Appendix_A}) \\
Optimizer & AdamW  \\
Learning Rate (LR) & 0.003  \\
Weight Decay (WD) & $1 \times 10^{-4}$  \\
Batch Size & 32  \\
Epochs per Generation & 50  \\
Number of Generations & 5  \\
Width Scaling ($\alpha$) & 0.5 -- 2.0  \\
Loss Function (ISIC/HeLa) & Mean Squared Error (MSE)  \\
Loss Function (SUIM/Cityscapes) & Categorical Cross-Entropy  \\
\midrule
Hardware & $2\times$ RTX 2080 Ti GPUs; Intel 9700k CPU; 64GB RAM \\ 
\bottomrule
\end{tabular}
\end{table}

\paragraph{Loss Function and Weighting.}
For both the HeLa and ISIC 2018 datasets, we utilized Mean Squared Error (MSE) on sigmoid-activated outputs.
We selected MSE after ablation studies where it yielded consistently better validation performance compared to Binary Cross-Entropy (including BCEWithLogits), Focal~\cite{lin_focal_2017} and Dice~\cite{sudre_generalised_2017} loss.

In the specific case of the multi-label HeLa dataset, MSE also outperformed hybrid combinations (e.g., BCE for masks combined with MSE for position regression).
Additionally, to compensate for the spatial sparsity of the cell-center targets, we applied a scalar weight of $3.0$ to the position-channel loss, while the \texttt{alive} and \texttt{dead} channels retained a weight of $1.0$.

\paragraph{Augmentation Protocols.}
Table~\ref{tab_augs_study_B} details the specific augmentation parameters used for Study B. We utilize Gaussian Blur, Gaussian Noise, and Brightness adjustments (controlled via Alpha and Beta parameters). The table lists the maximum intensity levels for five progressive stages. The application of these parameters varies by method:
\begin{itemize}
    \item \textbf{Progressive Methods (Noisy Student, IM+, IM++, AIM+, AIM++):} These methods ramp up augmentation intensity across generations. Generation 1 utilizes the first value in the list, incrementally increasing to the fifth value by Generation 5.
    \item \textbf{Maximum Intensity Baselines (FixMatch, FPL, iMAS, U$^2$PL, CrossMatch):} These methods utilize the \textbf{maximum intensity settings} (Stage 5) throughout their training to ensure they benefit from strong perturbations as defined in their respective protocols.
    \item \textbf{Moderate Intensity SSL (UniMatch):} Consistent with its specific protocol for these domains, UniMatch utilizes \textbf{moderate intensity settings} (corresponding roughly to Stage 3). We restricted this method to moderate levels after preliminary experiments indicated that maximum intensity augmentations degraded its performance.
    \item \textbf{Strong Supervised Baselines (ALDT, AFDT):} These models are trained using strong static augmentations (corresponding to Stage 5) to simulate a competitive supervised upper bound.
    \item \textbf{Static Methods (Standard IM, Ensembles):} These methods do not utilize any augmentations, serving as a control for the impact of data filtering versus data augmentation.
    \item \textbf{Internal Uncertainty SSL (UPS):} UPS does not utilize the photometric input augmentations listed in this table. We omitted input perturbations after preliminary experiments indicated that even mild augmentations destabilized the pseudo-label selection process in this low-data regime, further degrading performance. Instead, it relies exclusively on Monte Carlo Dropout within the network architecture to estimate uncertainty.
\end{itemize}

\begin{table*}[t]
\centering
\caption{Augmentation Schedule for Study B. Values represent the maximum intensity levels for each augmentation type across the 5 progressive stages. ``Brightness Alpha'' and ``Brightness Beta'' represent the $\pm$ range of random deviations.}
\label{tab_augs_study_B}
\resizebox{\textwidth}{!}{%
\begin{tabular}{@{}lllll@{}}
\toprule
\textbf{Dataset} & \textbf{Max Blur} & \textbf{Max Noise} & \textbf{Brightness Alpha ($\pm$)} & \textbf{Brightness Beta ($\pm$)}\\
& \textit{(Gen 1 $\rightarrow$ 5)} & \textit{(Gen 1 $\rightarrow$ 5)} & \textit{(Gen 1 $\rightarrow$ 5)} & \textit{(Gen 1 $\rightarrow$ 5)} \\
\midrule
ISIC 2018 & 0, 1, 1, 2, 3 & 5, 10, 15, 20, 25 & 0.1, 0.2, 0.3, 0.4, 0.5 & 5, 10, 15, 20, 25  \\
HeLa & 0, 1, 1, 2, 3 & 5, 10, 15, 20, 25 & 0.1, 0.1, 0.2, 0.2, 0.3 & 3, 6, 9, 12, 15 \\
SUIM & 0, 1, 1, 2, 3 & 5, 10, 15, 20, 25 & 0.1, 0.2, 0.3, 0.4, 0.5 & 5, 10, 15, 20, 25 \\
Cityscapes & 0, 0, 0, 0, 1 & 5, 10, 15, 20, 25 & 0.1, 0.1, 0.2, 0.2, 0.3 & 3, 6, 9, 12, 15 \\
\bottomrule
\end{tabular}%
}
\end{table*}

\section{Methodological Details for Study B} \label{Appendix_B}

In our resource-constrained study (Section~\ref{sec_study_b_results}), we compared Inconsistency Masks against a rigorous set of baselines representing established SSL paradigms.

\subsection{Standard SSL Baselines}

\paragraph{Iterative Self-Training (Noisy Student)} \label{noisy_student} We implemented the standard generational Noisy Student paradigm~\cite{xie_self-training_2019}. A teacher model generates pseudo-labels for a student, which is then trained with strong augmentations. Consistent with the original method, we increased the student model size (width factor $\alpha$) across generations. 

\paragraph{Consistency Loss (CL)} \label{consistency_loss}
We tested a simple consistency loss approach where a single model is trained to minimize the Mean Squared Error (MSE) between predictions on two symmetrically augmented views of the same unlabeled input. Unlike FixMatch, which employs asymmetric augmentation (weak-to-strong) and strict confidence thresholding to filter noise, this approach enforces consistency on all pixels regardless of prediction confidence. This lack of filtering forces the model to align predictions even on ambiguous regions, potentially reinforcing low-confidence errors.

\paragraph{EvalNet:} \label{evalnet}
We trained the EvalNet regression model (detailed in Appendix~\ref{Appendix_A}) to predict the IoU quality of pseudo-labels. We evaluated using EvalNet as a standalone filter by discarding pseudo-labels with predicted scores below a threshold derived from the ALDT baseline. This module is subsequently integrated as a secondary quality gate within the IM++ framework.

\subsection{Ensemble Baselines}
To isolate the specific contribution of the \textit{masking} mechanism, we investigated standard ensemble methods.

\paragraph{Model Ensemble (ME).} \label{baseline:ME} 
Pseudo-labels are generated via ensemble consensus (averaging probabilities for multi-class, strict intersection for binary) from independently trained teachers. The critical difference is that ME forces a specific prediction for every pixel based on the ensemble's aggregate confidence, whereas IM explicitly removes pixels where the teachers disagree.

\paragraph{Input Ensemble (IE).} \label{baseline:IE} 
Pseudo-labels are generated by a single teacher making predictions on multiple augmented views of the input. Similar to ME, these predictions are aggregated (via averaging or intersection) to form the pseudo-label, retaining all pixels regardless of disagreement levels.

\paragraph{Comparison of Baselines.} Our results (see Appendix~\ref{fig_results_Study_B}, ~\ref{fig_full_results_Study_B} and Table~\ref{tab_Study_B}) reveal a sharp distinction. While Input Ensembles (IE) performed poorly---often degrading below the baseline---Model Ensembles (ME) did yield improvements. However, ME consistently underperformed compared to IM (e.g., $37.1\%$ vs $44.3\%$ on SUIM). This confirms that simply averaging out disagreement is inferior to the IM strategy of explicitly filtering it out.

\subsection{Advanced Framework Variants}\label{Appendix_B_IM_Advanced_Variants}

While the core Inconsistency Mask (IM) framework effectively filters label noise, we explored two advanced extensions designed to further maximize performance in the resource-constrained regime (Study B). These variants, termed IM+ and IM++, integrate principles of curriculum learning and quality estimation to prevent the student model from plateauing.

\subsubsection{IM+: Progressive Learning}
The standard IM approach trains a new student from scratch at every generation using a fixed model architecture and a fixed set of data augmentations. However, as the quality of the teacher ensemble improves across generations, the student model faces diminishing returns if the learning task remains static. To address this, IM+ adopts the ``Noisy Student'' paradigm~\cite{xie_self-training_2019}, introducing progressive scaling in both model capacity and data perturbation.

\vspace{0.5em} \noindent \textbf{Dynamic Architecture Scaling:} We utilize the flexible design of our $1\times1$ U-Net (see Appendix~\ref{Appendix_A}) to scale the network width across generations. In the initial generation (Gen 0), we train a compact model with a width factor of $\alpha=0.5$. In subsequent generations, as the pseudo-labels become more reliable, we increase $\alpha$ (e.g., to $1.0$, $1.5$, and $2.0$). This ensures that the student model has sufficient capacity to learn increasingly complex features distilled from the teacher ensemble.

\vspace{0.5em} \noindent \textbf{Progressive Augmentation:} Similarly, we ramp up the intensity of data augmentations. Early generations utilize weak augmentations (e.g., simple flips and rotations) to ensure stability. Later generations introduce strong photometric distortions (color jitter, Gaussian noise) and geometric deformations. This forces the student to learn more robust invariant features, preventing it from merely memorizing the teacher's predictions.

\subsubsection{IM++: Quality Gating via EvalNet}
While Inconsistency Masks effectively filter local (pixel-level) uncertainty, they do not assess the global quality of a generated pseudo-label. A prediction might be consistent across the ensemble (consensus) but structurally incorrect (e.g., a ``blobby'' segmentation that misses fine details). IM++ addresses this by leveraging the  \hyperref[evalnet]{\textbf{EvalNet}} module for explicit quality assessment.

\vspace{0.5em} \noindent \textbf{Quality Estimation:} The EvalNet (architecture detailed in Appendix~\ref{Appendix_A_EvalNet}) is a lightweight regression network trained to predict the Intersection-over-Union (IoU) between an image and its segmentation mask. During the pseudo-labeling phase, we pass the unlabeled image and its ensemble-generated mask through EvalNet to obtain a predicted quality score $q \in [0, 1]$.

\vspace{0.5em} \noindent \textbf{Dynamic Data Generation:} Instead of a binary inclusion/exclusion threshold, we use $q$ to modulate the training frequency of each sample. We employ a dynamic generation strategy where the number of augmented copies created for an image $x_i$ is proportional to its quality score $q_i$. High-quality pseudo-labels generate multiple augmented training samples, acting as strong anchors, while low-quality labels (which might pass the IM filter but look structurally dubious) generate fewer samples. This focuses the computational budget on the most reliable data, allowing IM++ to surpass the standard FDT reference in some settings.

\subsubsection{Augmented Variants (AIM+ / AIM++)}
To push the performance ceiling further, we introduce Augmented IM+ (AIM+) and AIM++. These variants follow the logic of IM+ and IM++ respectively but utilize the Augmented Labeled Dataset (ALD) instead of the standard labeled set for the supervised loss component. This allows the model to leverage strong augmentations on the labeled data alongside the filtered pseudo-labels.

\subsection{Handling of Inconsistent Pixels in Study B}
\label{sec:handling_inconsistent_pixels}

For the experiments in Study B, we utilized an ``Explicit Uncertainty Class'' strategy. We shifted the original dataset class indices by $+1$ and assigned index $0$ to a new ``Inconsistency'' class. During training, pixels identified by the Inconsistency Mask were explicitly labeled as Class $0$.

This design choice was specifically adopted to enhance robustness in the ``training from scratch'' regime. Unlike a standard \texttt{ignore\_index}, which allows the model to predict any class without penalty in masked regions, this strategy explicitly \textbf{suppresses the probability mass} for semantic classes (e.g., cars, pedestrians) in discordant regions. This forces the student model to be conservative, preventing it from hallucinating specific class features in ambiguous areas where the teacher ensemble could not agree. This active suppression of ambiguity contributes directly to the training stability observed in this regime.

\paragraph{Fairness of Comparison.} To ensure a fair comparison, all baseline methods were trained on the exact same shifted datasets using the same network architecture with $C+1$ output channels. Since the baseline methods never encounter Class $0$ in the labeled data (and mask out low-confidence predictions on unlabeled data), the output neuron for Class $0$ effectively becomes inactive for them. Thus, the performance of the baselines is not negatively impacted by the class shift, ensuring the validity of the comparison.

\paragraph{Quality Filtering for Binary Tasks.}
For the ISIC 2018 dataset (binary segmentation), we implemented an additional quality gate based on the relative size of the Inconsistency Mask. We observed that if the area of the IM exceeds the area of the predicted foreground (lesion), the pseudo-label is highly unreliable. Consequently, we discard such input-pseudo-label pairs from the training set entirely for the current generation. This heuristic serves a dual purpose: it prevents the student from training on incoherent structures and simultaneously accelerates the training process by reducing the effective dataset size, ensuring computational resources are focused only on informative samples.

\section{Dataset Specifications}

\subsection{Public Benchmark Datasets}
For Study B, we standardized the evaluation by partitioning all training sets into 10\% labeled and 90\% unlabeled splits. Exact sample counts for all partitions are detailed in Table~\ref{tab:dataset_stats}.

\paragraph{ISIC 2018} The dataset from the International Skin Imaging Collaboration~\cite{codella_skin_2019, tschandl_ham10000_2018} focuses on binary skin lesion segmentation. We utilized the 2,594 training images, resizing them uniformly to $256 \times 256$ pixels.

\paragraph{SUIM} The Segmentation of Underwater Imagery dataset~\cite{islam_semantic_2020} provides pixel-level annotations for 8 object categories. To standardized inputs for our architecture, we processed the images into $256 \times 256$ patches by extracting \textbf{2 random crops per image}.
\begin{itemize}
    \item \textbf{Test Set:} Generated from the official SUIM test set (resulting in 250 patches).
    \item \textbf{Validation Set:} Generated by splitting 10\% of the official SUIM training set (resulting in 250 patches).
    \item \textbf{Training Set:} Generated from the remaining 90\% of the official SUIM training set (resulting in 2,744 patches).
\end{itemize}

\paragraph{Cityscapes} For \textbf{Study A} (Pre-trained Benchmarks), we adhered to the standard protocols for Cityscapes, utilizing the official 2,975 training images and reporting results on the full 500-image validation set using the standard 19-class taxonomy.

For the resource-constrained experiments (\textbf{Study B}), we utilized the full 34-class taxonomy (including void classes) to test fine-grained discrimination. To create a local test environment without submitting to the evaluation server:
\begin{itemize}
    \item \textbf{Training:} We used the official 2,975 training images.
    \item \textbf{Validation \& Testing:} We randomly partitioned the official 500-image validation set into two equal subsets: a \textbf{Local Validation Set} (250 images) utilized during the training process for monitoring and checkpoint selection, and a \textbf{Local Test Set} (250 images) reserved exclusively for generating the final performance metrics reported in Tables~\ref{tab_Study_B}, \ref{tab_appendix_standard_metrics}, and \ref{tab_appendix_alternative_metrics}.
\end{itemize}
Inputs for Study B were downsampled to $208 \times 416$ pixels.

\begin{table*}[h]
\caption{Dataset statistics and split definitions for Study B (Resource-Constrained Regimes). 
\textbf{FD:} Full Dataset. \textbf{LD:} Labeled Subset (10\%). \textbf{ULD:} Unlabeled Subset (90\%). \textbf{ALD:} Augmented Labeled Dataset (used for the ALDT baseline). 
Input shapes are denoted as $H \times W \times C$.}
\label{tab:dataset_stats}
\centering
\begin{tabular*}{\textwidth}{@{\extracolsep{\fill}}lllllllll@{}}
\toprule
\makecell[l]{Dataset} & \makecell[l]{Shape\\($H \times W \times C$)} & \makecell[l]{FD} & \makecell[l]{LD} & \makecell[l]{ALD} & \makecell[l]{ULD} & \makecell[l]{Validation \\ Set} & \makecell[l]{Test \\ Set} & \makecell[l]{Number of \\ Classes} \\ 
\midrule
ISIC 2018  & $256\times256\times3$ & 2,594 & 259 & 2,590 & 2,332 & 100 & 1,000 & 1 \\
HeLa       & $256\times256\times1$ & 2,380 & 238 & 2,380 & 2,142 & 420 & 420   & 3 \\
SUIM       & $256\times256\times3$ & 2,744 & 276 & 2,760 & 2,468 & 250 & 250   & 8 \\
Cityscapes & $208\times416\times3$ & 2,975 & 297 & 2,970 & 2,678 & 250 & 250   & 34 \\
\bottomrule
\end{tabular*}
\end{table*}

\subsection{Custom Evaluation Metric: MCCE}
\label{sec_mcce_def}

For the HeLa dataset, the primary downstream utility is accurate quantification of viable versus non-viable cell populations.
Standard segmentation metrics such as mIoU or Dice can be misleading in this context; a model may achieve high spatial overlap on large cell bodies while missing smaller instances, or fail to distinguish between cell states.
To address this, we utilize the \textbf{Mean Cell Count Error (MCCE)} as a key evaluation metric.

MCCE measures the average absolute deviation between the predicted and ground-truth counts, summed across the relevant biological categories. It is defined as:
\begin{equation}
    \text{MCCE} = \frac{1}{N} \sum_{i=1}^{N} \sum_{t=1}^{T} \left| c_t^{(i)} - p_t^{(i)} \right|,
    \label{eq:mcce}
\end{equation}

where lower values indicate better counting accuracy. The variables are defined as follows:

\begin{itemize}
    \item $N$ is the total number of images in the test set.
    \item $T$ represents the distinct cell categories (here $T=2$ for \texttt{alive} and \texttt{dead} cells).
    \item $c_t^{(i)}$ is the ground truth count of cells of type $t$ in image $i$.
    \item $p_t^{(i)}$ is the predicted count of cells of type $t$ in image $i$, obtained by thresholding the predicted mask and counting connected components per class.
\end{itemize}

\section{Supplementary Results}

In this section, we present additional ablations and qualitative results to support the main findings. 

\paragraph{Note on Metric Scales.} Note on Metric Scales. All performance graphs in this appendix display metrics (e.g., IoU, mIoU) on the $[0, 1]$ scale. These values are directly equivalent to the percentage-based results ($0$--$100\%$) reported in the tables throughout the main text and Appendix.

\subsection{Impact of Morphology and Resolution} \label{Appendix_D_K_e_d}
\begin{figure}[h] 
   \centering 
   \includegraphics[width=1\textwidth]{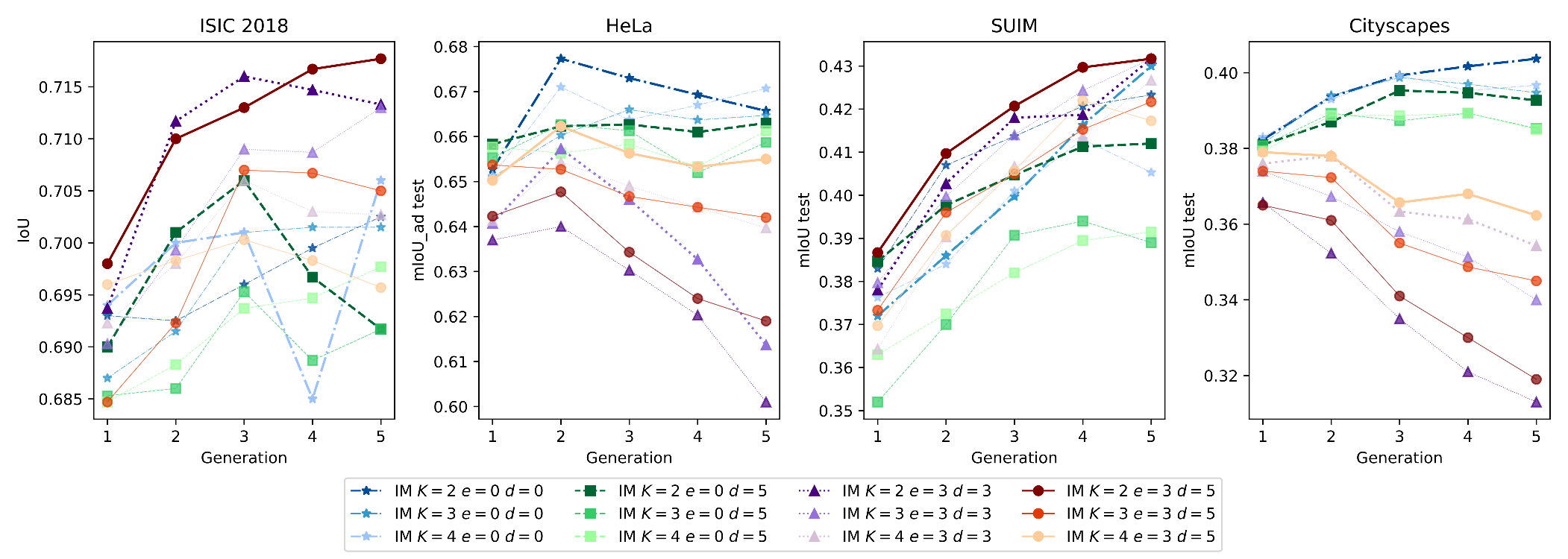} 
   \caption{Impact of Inconsistency Mask parameters on performance. $K$ represents the number of models used, $e$ the erode kernel size, and $d$ the dilate kernel size. The best results for each combination of $K$, $e$, and $d$ are highlighted in bold, while all others are displayed in a lighter, faded font.} 
   \label{fig_D_1} 
\end{figure}

Reviewing our parameter sweep (Fig.~\ref{fig_D_1}), we observed that aggressive morphological erosion ($e=3, d=5$) degraded performance on Cityscapes in our resource-constrained setup (Study B). We attribute this to the low input resolution ($208 \times 416$) used in Study B; at this scale, a $3 \times 3$ erosion kernel often eliminates entire fine-grained objects (e.g., poles, traffic signs).

However, for Study A (SOTA Benchmarks), we successfully utilized $e=3, d=5$. This was possible because the input resolution was significantly higher ($768 \times 768$). At this scale, erosion acts as intended---removing ambiguous boundary pixels to create a high-confidence ``core'' mask---without destroying the structural integrity of small objects. This suggests that the morphological parameters $e$ and $d$ should be adjusted according to the input resolution.

\textbf{Takeaway on $K$.} Fig.~\ref{fig_D_1} also evaluates the effect of the number of teacher models K. Across datasets and generations, we find that K=2 is a robust default: moving from K=2 to K=3 or K=4 yields only occasional gains and does not improve performance consistently. Since larger K increases the cost of pseudo-label generation (additional teacher forward passes) without reliable benefits, we use K=2 throughout the paper unless otherwise stated.

\subsection{Experimental Protocol \& Reliability}
To ensure robustness, all experiments were performed with three independent repetitions. The graphical analyses in this appendix visualize the \textbf{mean test set performance} to illustrate training dynamics and stability. In contrast, the tabular results (Tables~\ref{tab_appendix_standard_metrics} and \ref{tab_appendix_alternative_metrics}) report the \textbf{single best test result achieved} across these repetitions. This protocol maintains consistency with the peak performance metrics reported in the main text (Table~\ref{tab_Study_B}).

\subsection{Full Generalization Results (Study B)}
In this section, we present the comprehensive performance metrics for all evaluated methods across the four datasets. Tables~\ref{tab_appendix_standard_metrics} and ~\ref{tab_appendix_alternative_metrics} benchmark the proposed Inconsistency Mask variants (IM, IM+, IM++) against a rigorous suite of Reference (e.g., LDT, FDT), Internal (e.g., Ensembles, Noisy Student), and State-of-the-Art (e.g., FixMatch, iMAS) baselines, providing both standard (IoU) and alternative metrics.

\begin{table*}[t]
\centering
\caption{\textbf{Standard segmentation performance (IoU / mIoU in \%).} Results reflect the highest scores from all experimental repetitions. Bold values with gray background indicate the best SSL performance per dataset. UPS failed to run on Cityscapes due to high memory requirements.}
\label{tab_appendix_standard_metrics}
\resizebox{\textwidth}{!}{%
\begin{tabular}{lcccc}
\toprule
\textbf{Approach} & \textbf{ISIC 2018} & \textbf{HeLa} & \textbf{SUIM} & \textbf{Cityscapes} \\
 & (IoU $\uparrow$) & (mIoU\textsubscript{ad} $\uparrow$) & (mIoU $\uparrow$) & (mIoU $\uparrow$) \\
\midrule
\multicolumn{5}{l}{\textit{Reference Baselines}} \\
LDT & 67.1 & 56.5 & 35.7 & 32.0 \\
ALDT & 72.4 & 65.9 & 43.2 & 37.4 \\
FDT & 75.1 & 69.6 & 51.7 & 45.6 \\
AFDT & 77.3 & 71.3 & 52.7 & 45.8 \\
\midrule
\multicolumn{5}{l}{\textit{Internal Baselines}} \\
Model Ensemble (ME) & 69.0 & 59.2 & 37.1 & 35.0 \\
Input Ensemble (IE) & 68.1 & 36.9 & 34.0 & 26.2 \\
Consistency Loss (CL) & 58.8 & 60.1 & 27.4 & 34.5 \\
Noisy Student (NS) & 74.3 & 68.0 & 43.2 & 40.0 \\
EvalNet & 73.7 & 66.3 & 40.8 & 35.5 \\
\midrule
\multicolumn{5}{l}{\textit{SOTA Baselines}} \\
FixMatch~\cite{sohn2020fixmatch} & 70.3 & 63.1 & 36.1 & 36.6 \\
FPL~\cite{fuzzypositivlearning} & 68.4 & 64.4 & 25.7 & 15.2 \\
CrossMatch~\cite{zhao2024crossmatch} & 65.7 & 62.3 & 36.5 & 34.7 \\
iMAS~\cite{iMAS} & 66.1 & 61.1 & 33.7 & 35.2 \\
UPS~\cite{UPS} & 37.9 & 60.7 & 26.8 & -- \\
U$^2$PL~\cite{u2pl} & 67.5 & 61.6 & 36.6 & 35.5 \\
UniMatch~\cite{UniMatch} & 64.0 & 61.9 & 26.5 & 24.3 \\
\midrule
\multicolumn{5}{l}{\textit{Ours}} \\
IM & 72.3 & 68.2 & 44.3 & 40.7 \\
IM+ & 73.9 & 69.2 & 45.4 & 42.7 \\
IM++ & 76.2 & 68.6 & 46.8 & \cellcolor{bestgray}\textbf{42.8} \\
AIM+ & 75.3 & \cellcolor{bestgray}\textbf{69.5} & 48.2 & 41.2 \\
AIM++ & \cellcolor{bestgray}\textbf{77.0} & 69.4 & \cellcolor{bestgray}\textbf{48.5} & 42.4 \\
\bottomrule
\end{tabular}%
}
\end{table*}

\begin{table*}[t]
\centering
\caption{\textbf{Alternative segmentation performance.} Metrics are Dice Score (\%) and Mean Pixel Accuracy (mPA \%). For HeLa, the metric is Mean Cell Count Error (MCCE, absolute count), where lower is better. Bold values with gray background indicate the best SSL performance per dataset. UPS failed to run on Cityscapes due to high memory requirements.}
\label{tab_appendix_alternative_metrics}
\resizebox{\textwidth}{!}{%
\begin{tabular}{lcccc}
\toprule
\textbf{Approach} & \textbf{ISIC 2018} & \textbf{HeLa} & \textbf{SUIM} & \textbf{Cityscapes} \\
 & (Dice \% $\uparrow$) & (MCCE $\downarrow$) & (mPA \% $\uparrow$) & (mPA \% $\uparrow$) \\
\midrule
\multicolumn{5}{l}{\textit{Reference Baselines}} \\
LDT & 76.1 & 9.91 & 57.9 & 73.3 \\
ALDT & 81.4 & 3.23 & 62.2 & 79.6 \\
FDT & 83.5 & 2.50 & 70.8 & 84.7 \\
AFDT & 85.3 & 2.41 & 70.4 & 85.0 \\
\midrule
\multicolumn{5}{l}{\textit{Internal Baselines}} \\
Model Ensemble (ME) & 78.0 & 3.94 & 59.1 & 79.5 \\
Input Ensemble (IE) & 76.8 & 27.25 & 56.3 & 69.0 \\
Consistency Loss (CL) & 69.1 & 19.35 & 48.4 & 77.9 \\
Noisy Student (NS) & 82.7 & 2.67 & 62.8 & 81.8 \\
EvalNet & 80.6 & 3.06 & 60.7 & 79.9 \\
\midrule
\multicolumn{5}{l}{\textit{SOTA Baselines}} \\
FixMatch~\cite{sohn2020fixmatch} & 79.0 & 42.60 & 59.0 & 80.6 \\
FPL~\cite{fuzzypositivlearning} & 77.1 & 30.60 & 47.2 & 62.5 \\
CrossMatch~\cite{zhao2024crossmatch} & 74.9 & 3.60 & 59.1 & 80.3 \\
iMAS~\cite{iMAS} & 76.7 & 13.80 & 58.1 & 79.8 \\
UPS~\cite{UPS} & 49.1 & 24.90 & 50.4 & -- \\
U$^2$PL~\cite{u2pl} & 77.0 & 22.60 & 58.0 & 79.6 \\
UniMatch~\cite{UniMatch} & 73.7 & 7.70 & 49.6 & 71.7 \\
\midrule
\multicolumn{5}{l}{\textit{Ours}} \\
IM & 80.7 & 2.77 & 66.4 & 82.3 \\
IM+ & 81.9 & 2.52 & 65.4 & 83.4 \\
IM++ & 84.0 & 2.51 & 66.0 & \cellcolor{bestgray}\textbf{83.7} \\
AIM+ & 83.3 & 2.50 & 67.3 & 82.9 \\
AIM++ & \cellcolor{bestgray}\textbf{85.0} & \cellcolor{bestgray}\textbf{2.49} & \cellcolor{bestgray}\textbf{67.8} & 83.2 \\
\bottomrule
\end{tabular}%
}
\end{table*}

\begin{figure}[b] 
   \centering 
   \includegraphics[width=1\textwidth]{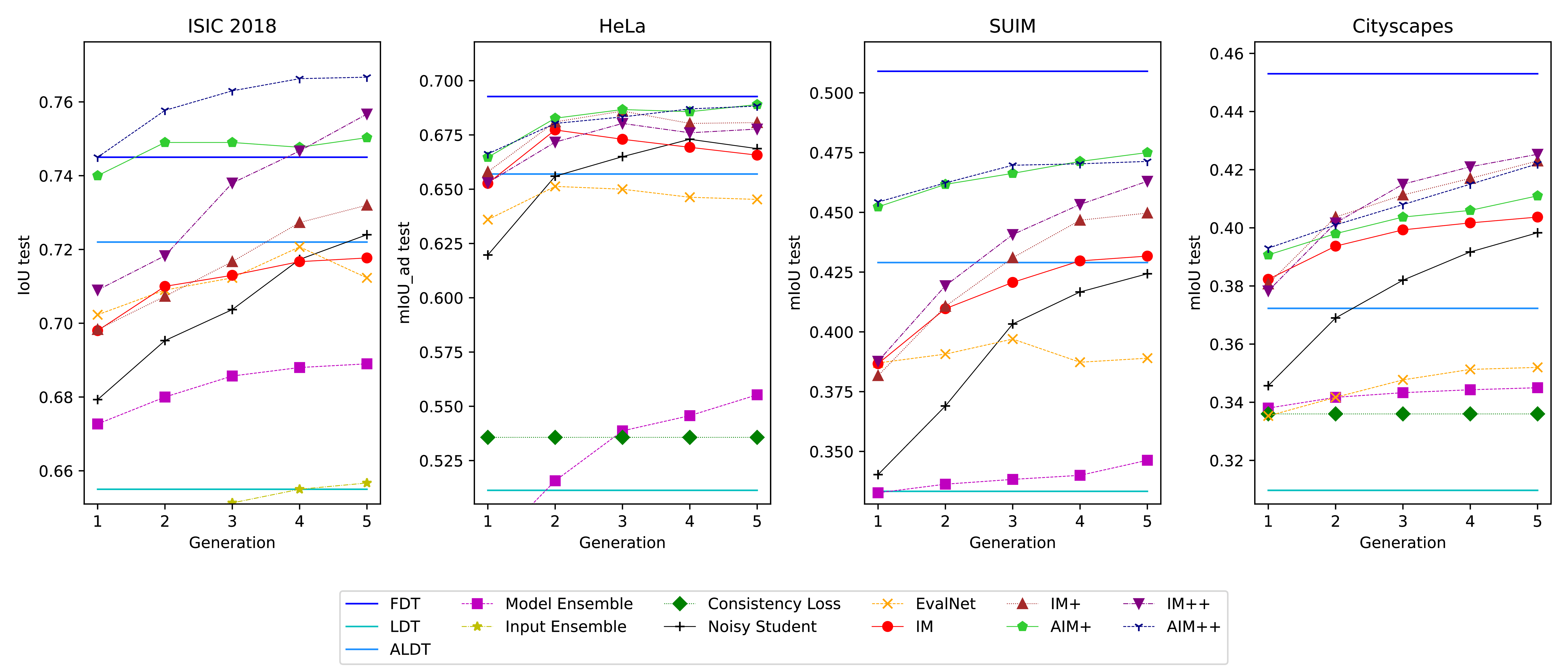} 
   \caption{presents the mean IoU/mIoU scores for all methods that outperformed the baseline, which is defined as training with the labeled subset (LD). Each experiment was repeated three times and benchmarked on the test sets to calculate the mean. For the HeLa dataset mIoU\_ad indicates that this is only the mIoU score for the two classes alive and dead.
   } 
   \label{fig_results_Study_B} 
\end{figure}

\begin{figure}[h] 
   \centering 
   \includegraphics[width=1\textwidth]{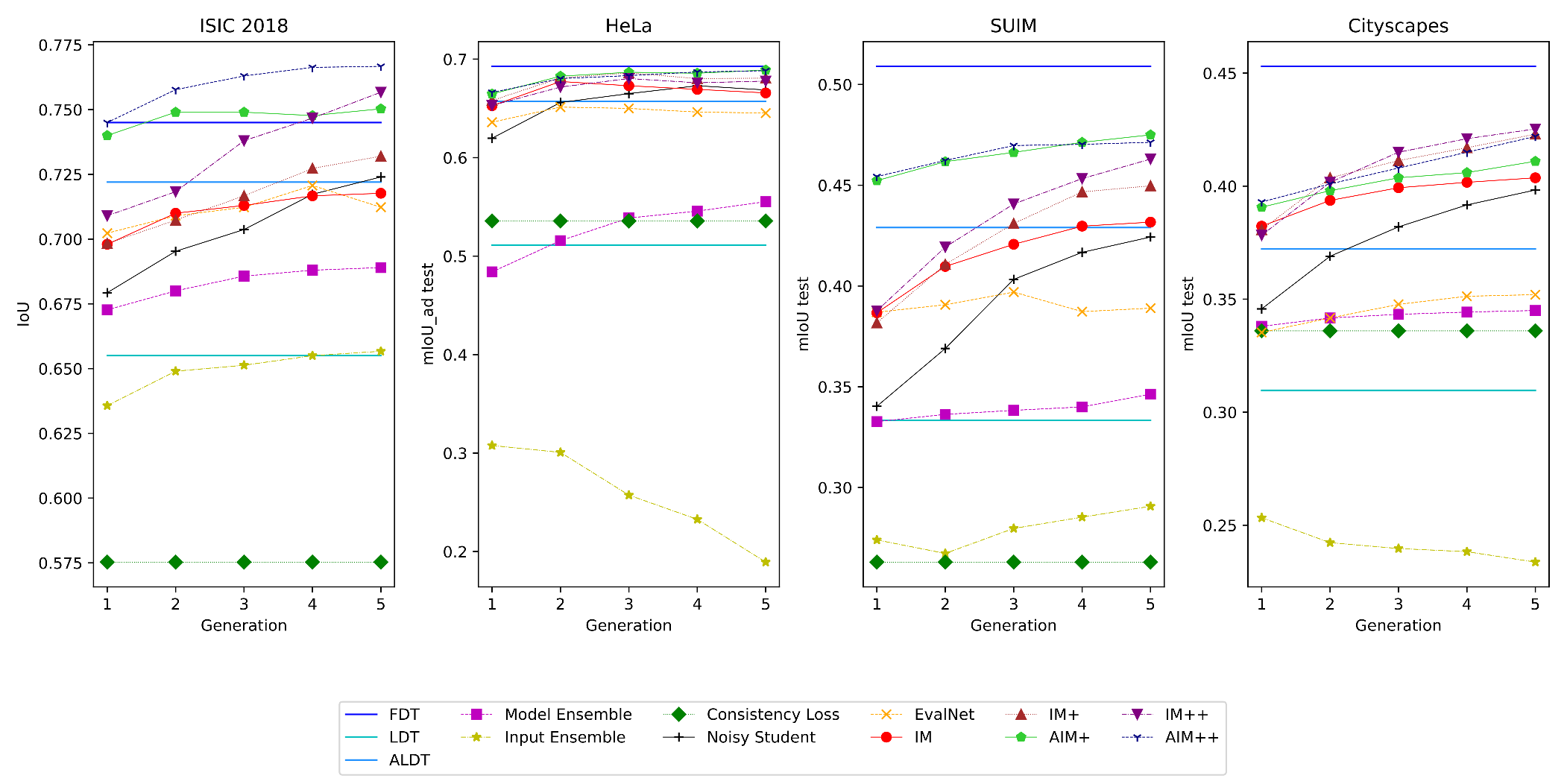} 
   \caption{Main Figure with all IoU / mIoU results.} 
   \label{fig_full_results_Study_B} 
\end{figure}

\begin{figure}[h] 
   \centering 
   \includegraphics[width=1\textwidth]{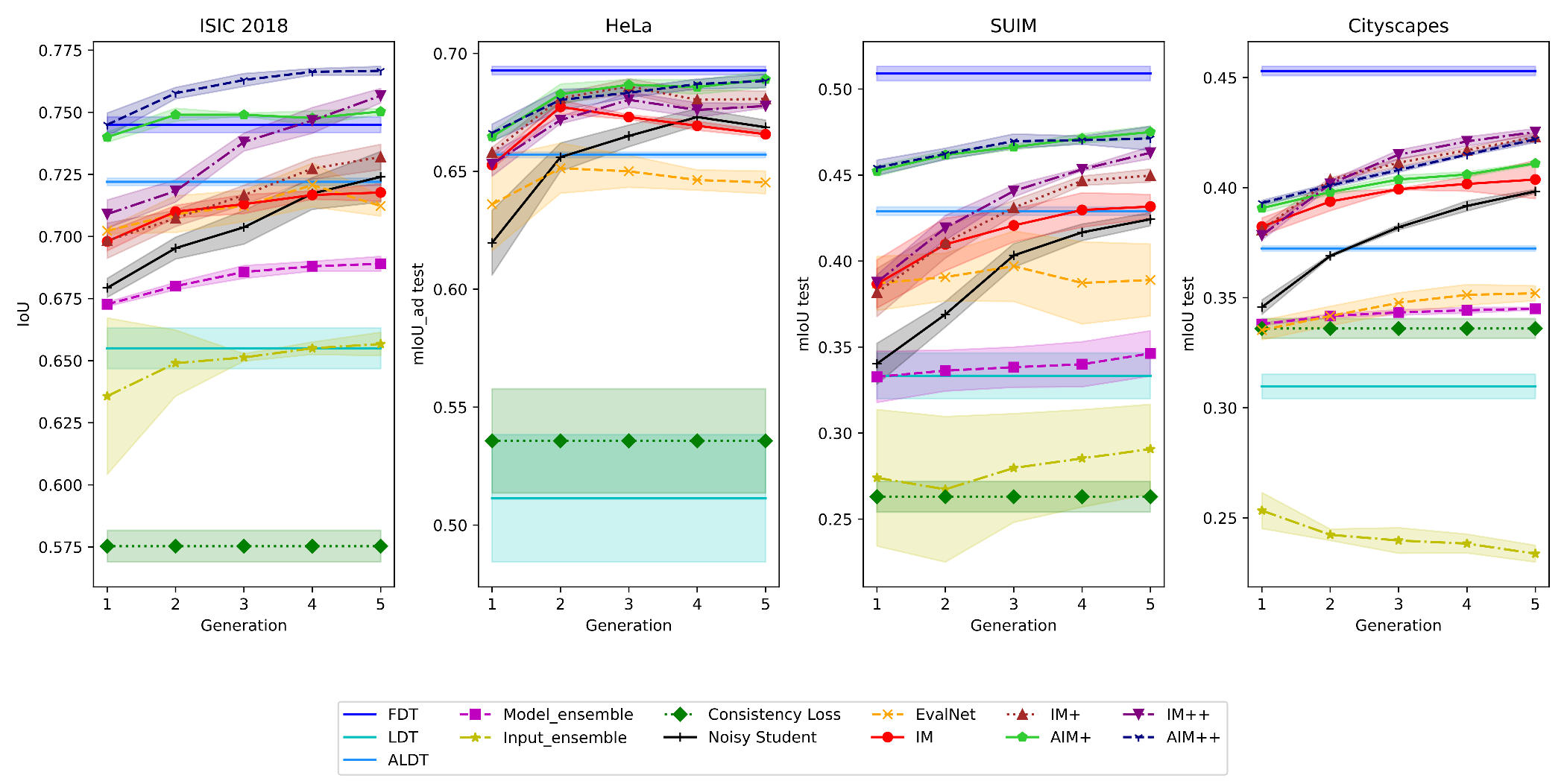} 
   \caption{Main results with standard error.} 
   \label{fig_full_results_with_std_err_Study_B} 
\end{figure}

\FloatBarrier
\subsection{Alternative Metrics}

\begin{figure}[!h] 
   \centering 
   \includegraphics[width=1\textwidth]{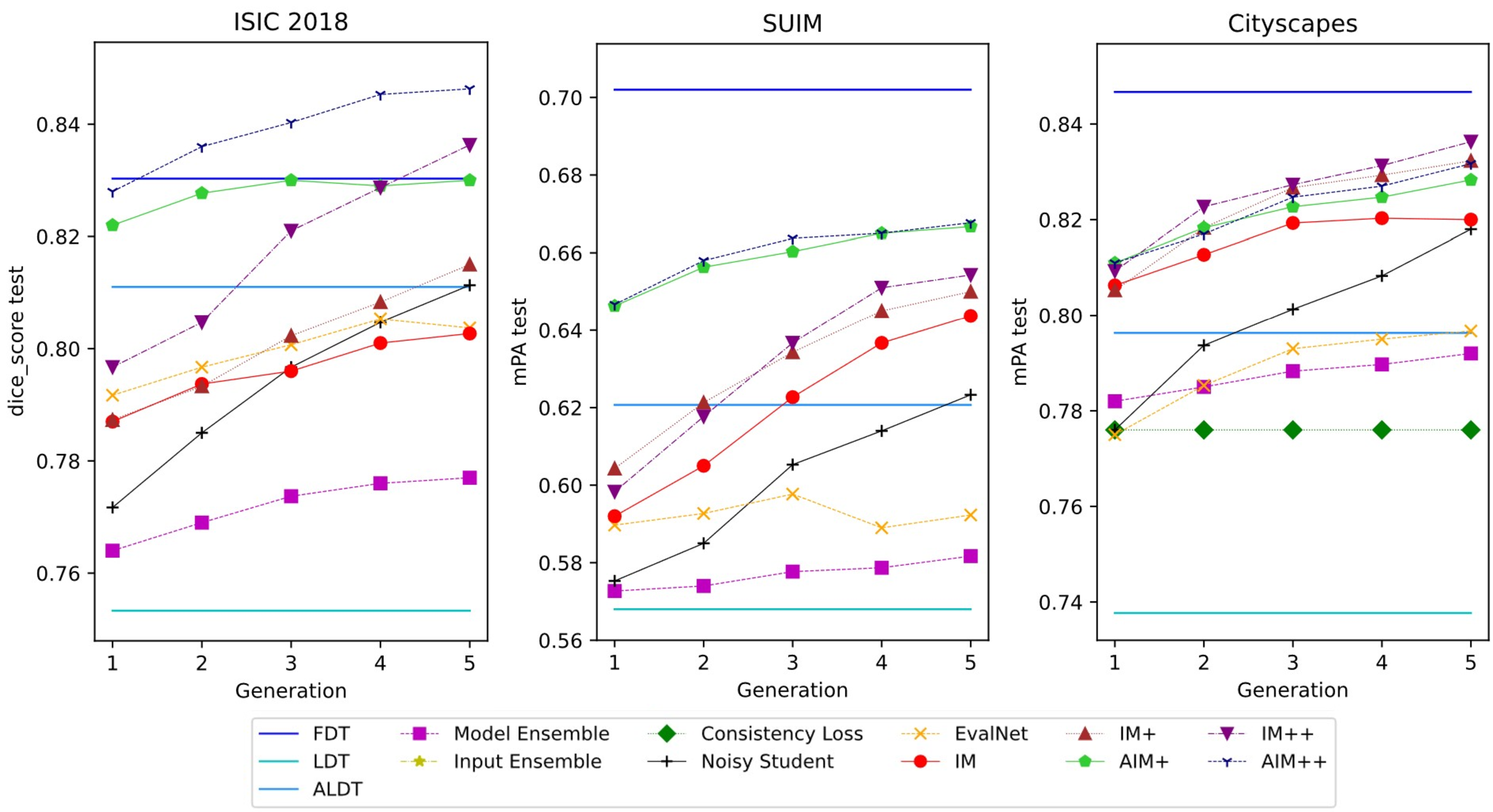} 
   \caption{Results on the test sets with alternative metrics. ISIC 2018 with dice score, SUIM and Cityscapes with mPA.} 
   \label{fig_full_results_alt_metrics_Study_B} 

   \vspace{0.5cm}
   
   \centering 
   \includegraphics[width=1\textwidth]{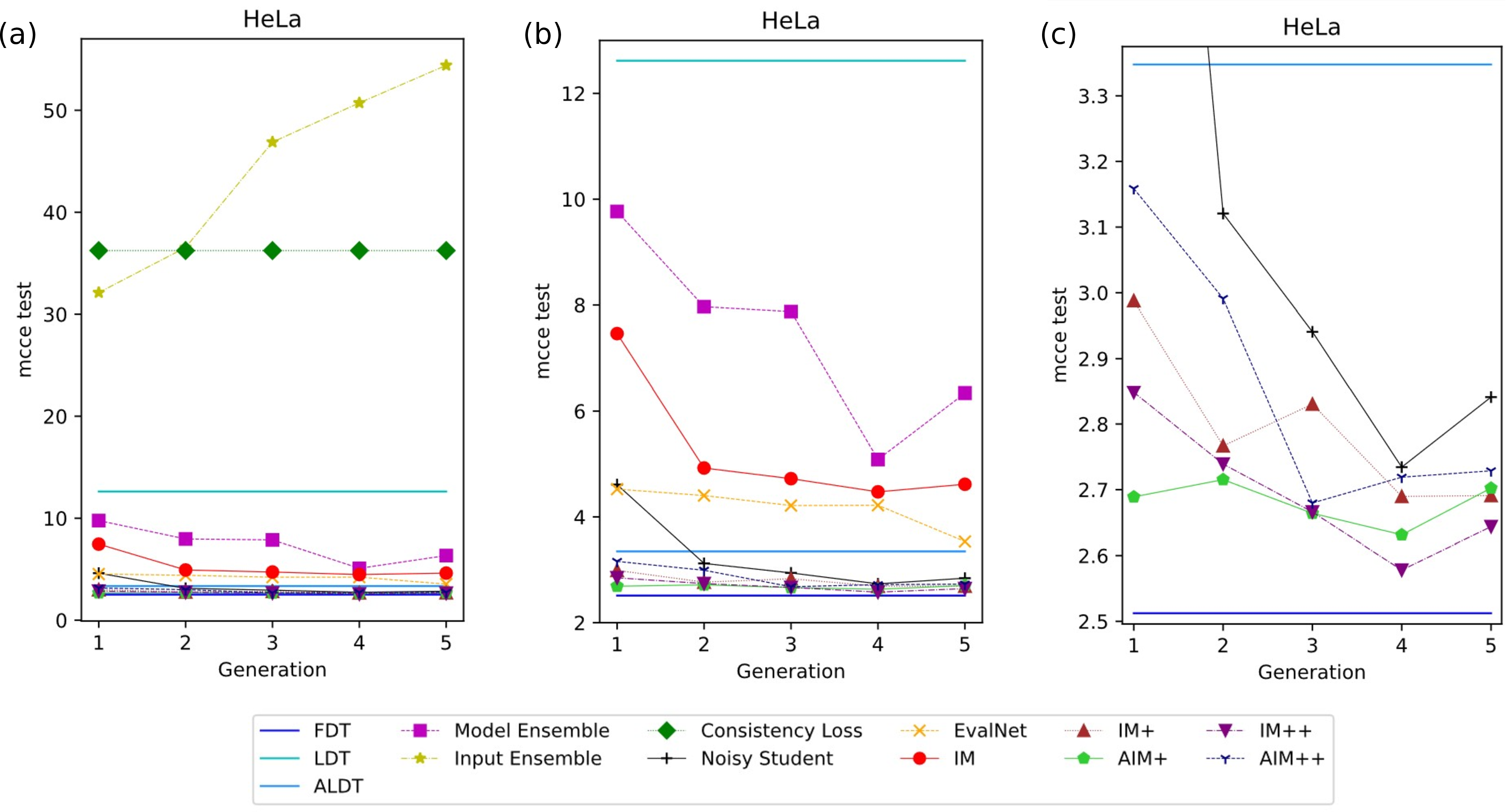} 
   \caption{MCCE results for the HeLa Dataset, lower is better. (a) shows all results, (b) zooms in to the range between LDT and FDT, (c) shows the range between ALDT and FDT.} 
   \label{fig_full_results_HeLa_Study_B} 
\end{figure}

\clearpage
\subsection{Ablation Studies}

\begin{figure*}[h] 
   \centering 
   \includegraphics[width=1\textwidth]{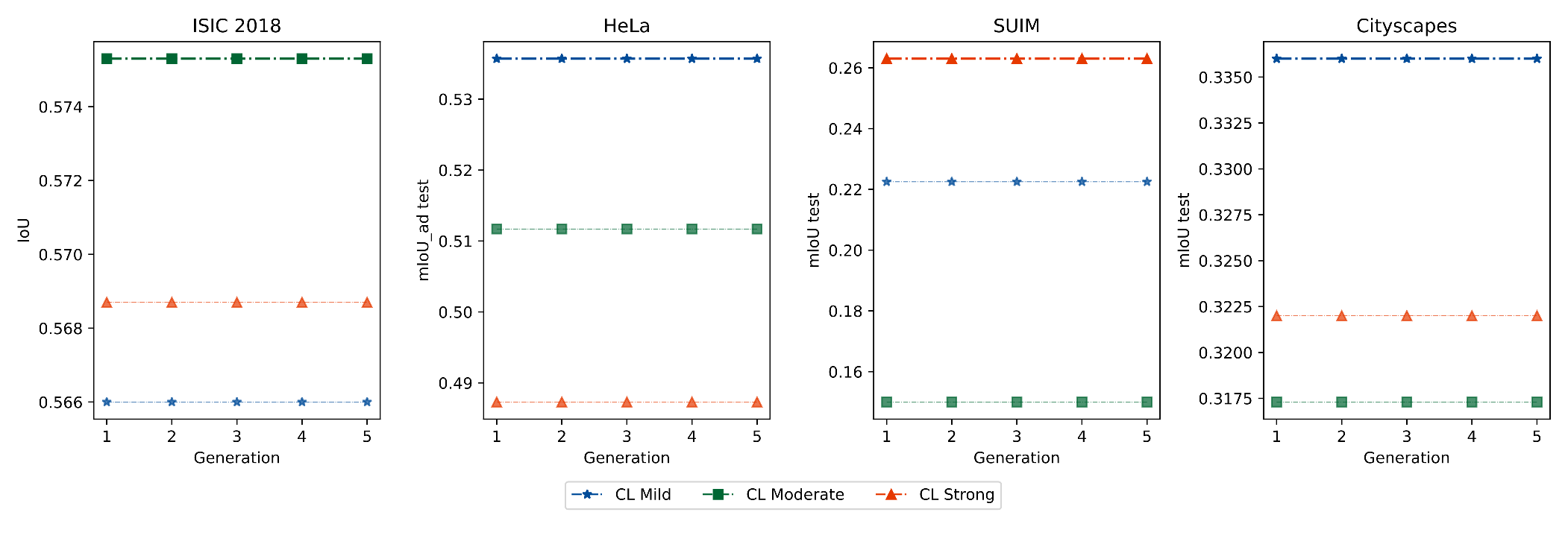} 
   \caption{Results of all Consistency Loss (CL) experiments using mild, moderate, and strong augmentations to produce the two augmented images for computing the consistency loss. The best results are highlighted in bold, while all others are displayed in a lighter, faded font.} 
   \label{fig19} 
\end{figure*}
    
\begin{figure*}[h] 
   \centering 
   \includegraphics[width=1\textwidth]{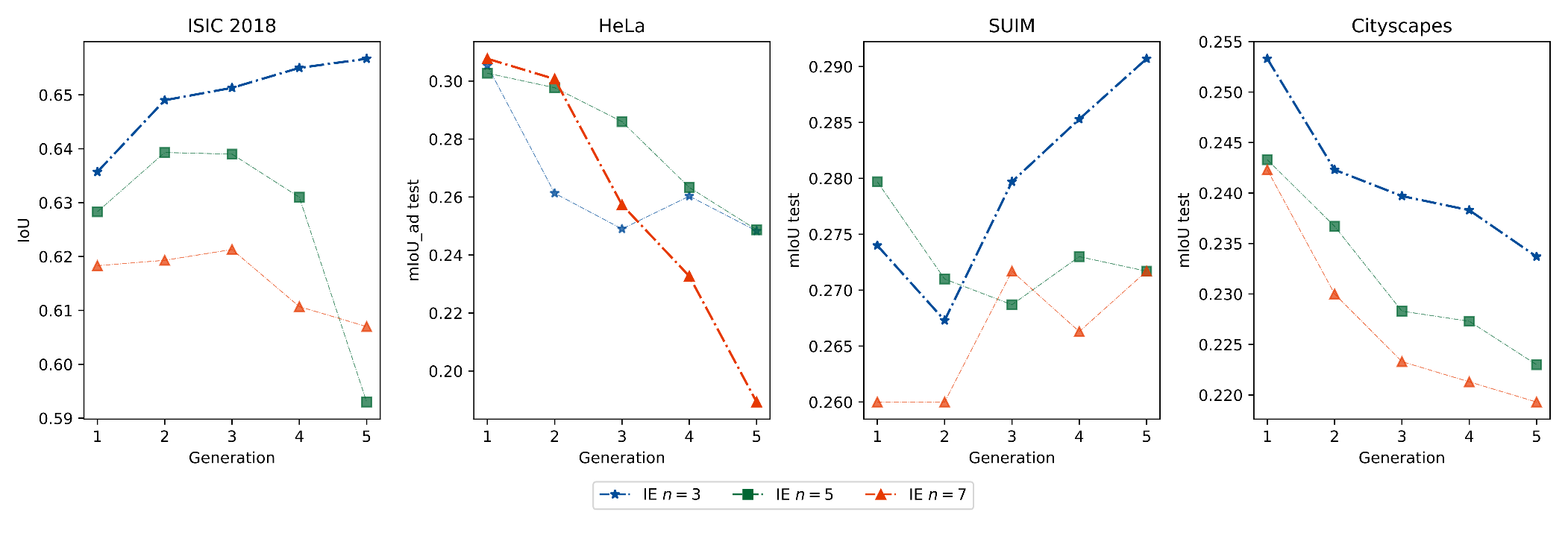} 
   \caption{Results of all trained Input Ensembles (IE). $n$ denotes the number of images used in each ensemble. The best results are highlighted in bold, while all others are displayed in a lighter, faded font.} 
   \label{fig20} 
\end{figure*}

\begin{figure*}[h] 
   \centering 
   \includegraphics[width=1\textwidth]{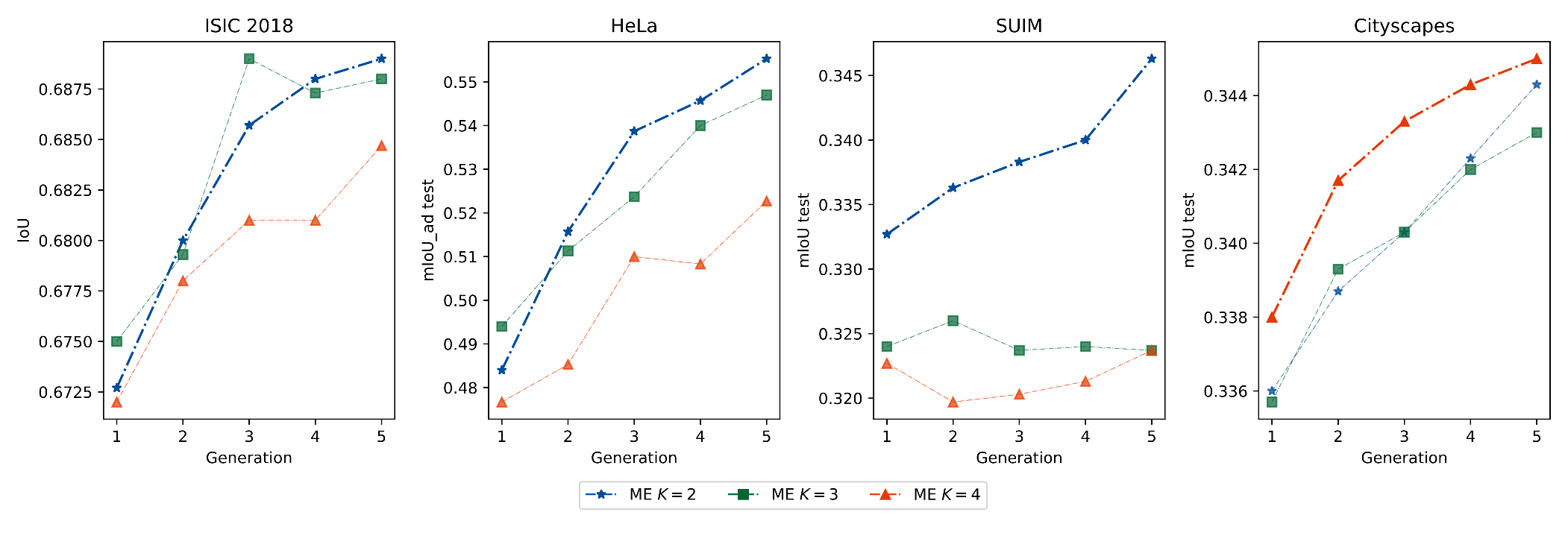} 
   \caption{Results of all trained Model Ensembles (ME). $K$ denotes the number of models used in each ensemble. The best results are highlighted in bold, while all others are displayed in a lighter, faded font.} 
   \label{fig21} 
\end{figure*}

\begin{figure*}[h] 
   \centering 
   \includegraphics[width=1\textwidth]{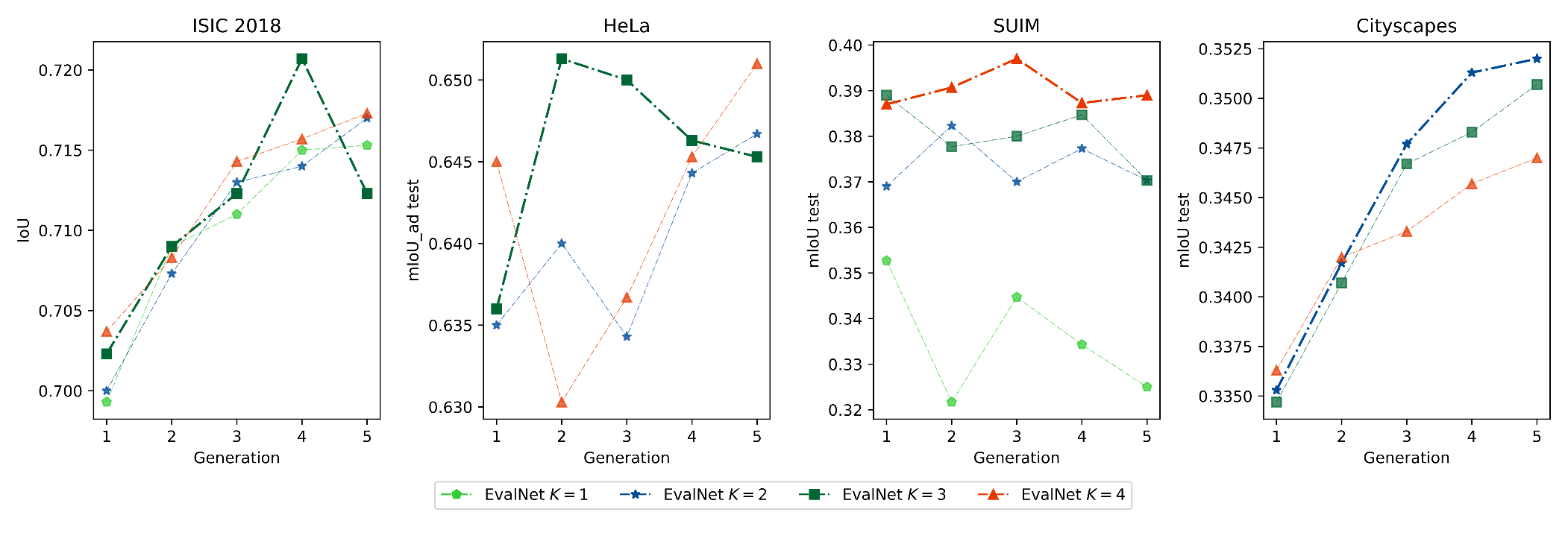} 
   \caption{Outcomes from all tested EvalNets and EvalNet Ensembles. $K=1$ denotes a single EvalNet, while $K>1$ signifies the ensemble size utilized. Due to the underperformance of single EvalNets in initial trials, we limited their training to ISIC 2018 and SUIM to verify the consistency of these preliminary findings. For ISIC 2018, the performance of the single EvalNet is remarkably close to that of the ensemble. However, for SUIM, the individual model displays significantly poorer results compared to the ensemble. Due to the clear performance advantage of EvalNet ensembles, further experiments with single EvalNets were not pursued.} 
   \label{fig22} 
\end{figure*}

\clearpage
\subsection{Qualitative Visualizations}
\begin{figure*}[h] 
   \centering 
   \includegraphics[width=1\textwidth]{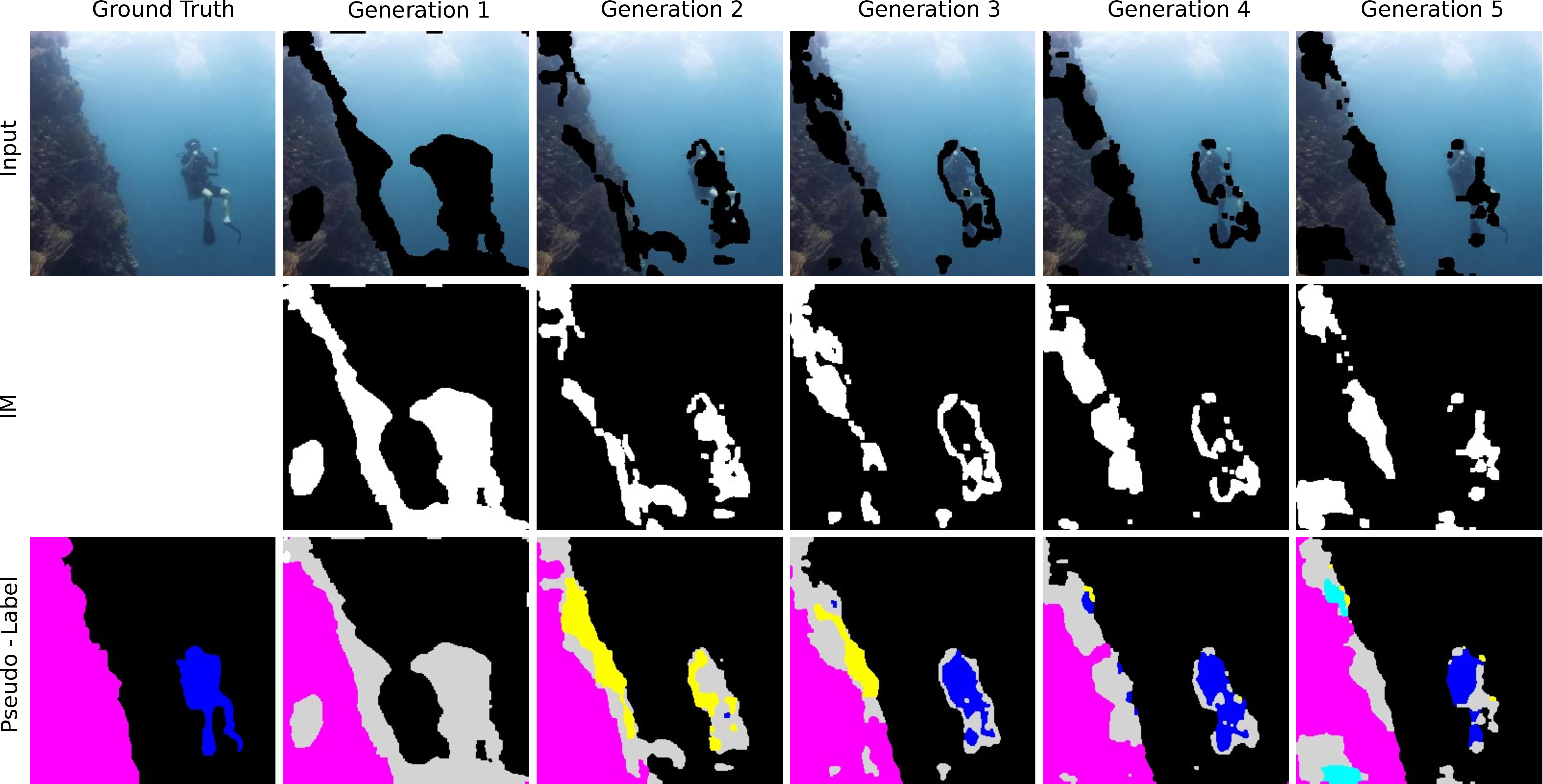} 
   \caption{Changes in input images and pseudo-labels on the SUIM dataset for IM+ over all five Generations. Magenta
represents reefs, black indicates background/waterbody , IM in gray, blue for divers, yellow for fish, and turquoise for wrecks.} 
   \label{fig_IM_quality_visualization} 
\end{figure*}

The size of the IM can, at least partially, serve as an indicator of the quality of the segmentation mask. For instance, filtering out input-pseudo-label pairs where the IM was larger than the sum of foreground pixels improved performance for the ISIC 2018 dataset. On the other hand, the pseudo-label mask from Generation 1, as illustrated in Fig. \ref{fig_IM_quality_visualization}, shows the highest number of correctly classified visible pixels of all the pseudo-label masks. Crucially, the IM successfully masks the gradient where the reef fades into the dark water—a naturally ambiguous region where defining a hard boundary is difficult even for humans. In the second Generation, although the IM is significantly smaller and more of the reef and diver becomes visible, the visible part of the diver and the newly visible part of the reef are mistakenly classified as fish. It is only in the subsequent Generation that the diver is correctly identified, and not until the following Generation that the visible part of the reef is almost entirely correctly classified.

\end{appendices}

\clearpage
\printbibliography


\end{document}